\documentclass[10pt,twocolumn,letterpaper]{article}

\usepackage{cvpr}
\usepackage{times}
\usepackage{epsfig}
\usepackage{graphicx}
\usepackage{amsmath}
\usepackage{amssymb}
\usepackage{color}
\usepackage{bm}
\usepackage{booktabs}
\usepackage{tablefootnote}
\usepackage{epigraph}
\usepackage{multirow}
\usepackage[caption=false]{subfig}
\usepackage{color}
\usepackage[table]{xcolor}
\usepackage{pdfpages}
\definecolor{mygray}{gray}{0.9}
\newcommand\ies{\textit{i.e.}}
\newcommand\egs{\textit{e.g.}}

\usepackage[pagebackref=true,breaklinks=true,letterpaper=true,colorlinks,bookmarks=false]{hyperref}

\cvprfinalcopy 

\def\cvprPaperID{} 

\begin{document}

\title{Visual Commonsense R-CNN}
\author{Tan Wang$^
{1,3}$, 
\quad Jianqiang Huang$^{2,3}$, 
\quad Hanwang Zhang$^3$, 
\quad Qianru Sun$^4$\\
\small $^1$University of Electronic Science and Technology of China~~
\small $^2$Damo Academy, Alibaba Group\\
\small $^3$Nanyang Technological University~~
\small $^4$Singapore Management University\\
\small {\texttt{wangt97@hotmail.com}},~
\small {\texttt{jianqiang.jqh@gmail.com}},~
\small {\texttt{hanwangzhang@ntu.edu.sg}},~
\small {\texttt{qianrusun@smu.edu.sg}}
\\
}

\maketitle

\begin{abstract}
We present a novel unsupervised feature representation learning method, Visual Commonsense Region-based Convolutional Neural Network (\textbf{VC R-CNN}), to serve as an improved visual region encoder for high-level tasks such as captioning and VQA. Given a set of detected object regions in an image (e.g., using Faster R-CNN), like any other unsupervised feature learning methods (e.g., word2vec), the proxy training objective of VC R-CNN is to predict the contextual objects of a region. However, they are fundamentally different: the prediction of VC R-CNN is by using \textbf{causal intervention}: $P(Y|do(X))$, while others are by using the conventional \textbf{likelihood}: $P(Y|X)$. This is also the core reason why VC R-CNN can learn ``sense-making'' knowledge 
like \texttt{chair} can be sat --- while not just ``common'' co-occurrences such as \texttt{chair} is likely to exist if \texttt{table} is observed. We extensively apply VC R-CNN features in prevailing models of three popular tasks: Image Captioning, VQA, and VCR, and observe consistent performance boosts across them, achieving many new state-of-the-arts\footnote{\url{https://github.com/Wangt-CN/VC-R-CNN}}.
\end{abstract}


\vspace{-0.1cm}
\section{Introduction}

\setlength{\epigraphwidth}{.9\columnwidth}
\renewcommand{\epigraphflush}{center}
\renewcommand{\textflush}{flushepinormal}
\epigraph{\textit{``On the contrary, Watson, you can see everything. You fail, however, to reason from what you see."} }
{{\footnotesize{\textit{--Sherlock Holmes, The Adventure of the Blue Carbuncle}}}}

\vspace{-0.13in}
Today's computer vision systems are good at telling us ``what'' (\eg, classification~\cite{he2016deep, krizhevsky2012imagenet}, segmentation~\cite{he2017mask, long2015fully}) and ``where'' (\eg, detection~\cite{ren2015faster, liu2016ssd}, tracking~\cite{kristan2015visual, li2019siamrpn++}), yet bad at knowing ``why'', \eg, why is it \texttt{dog}? Note that the ``why'' here does not merely mean by asking for \emph{visual} reasons --- attributes like furry and four-legged --- that are already well-addressed by machines; beyond, it also means by asking for high-level \emph{commonsense} reasons --- 
such as \texttt{dog} barks~\cite{gibson1977theory} --- that are still elusive, even for us human philosophers~\cite{rosenfeld2011common, halloun1985common,smith1995structures}, not to mention for machines.

\begin{figure}[t]
\begin{center}
\includegraphics[width=0.45\textwidth]{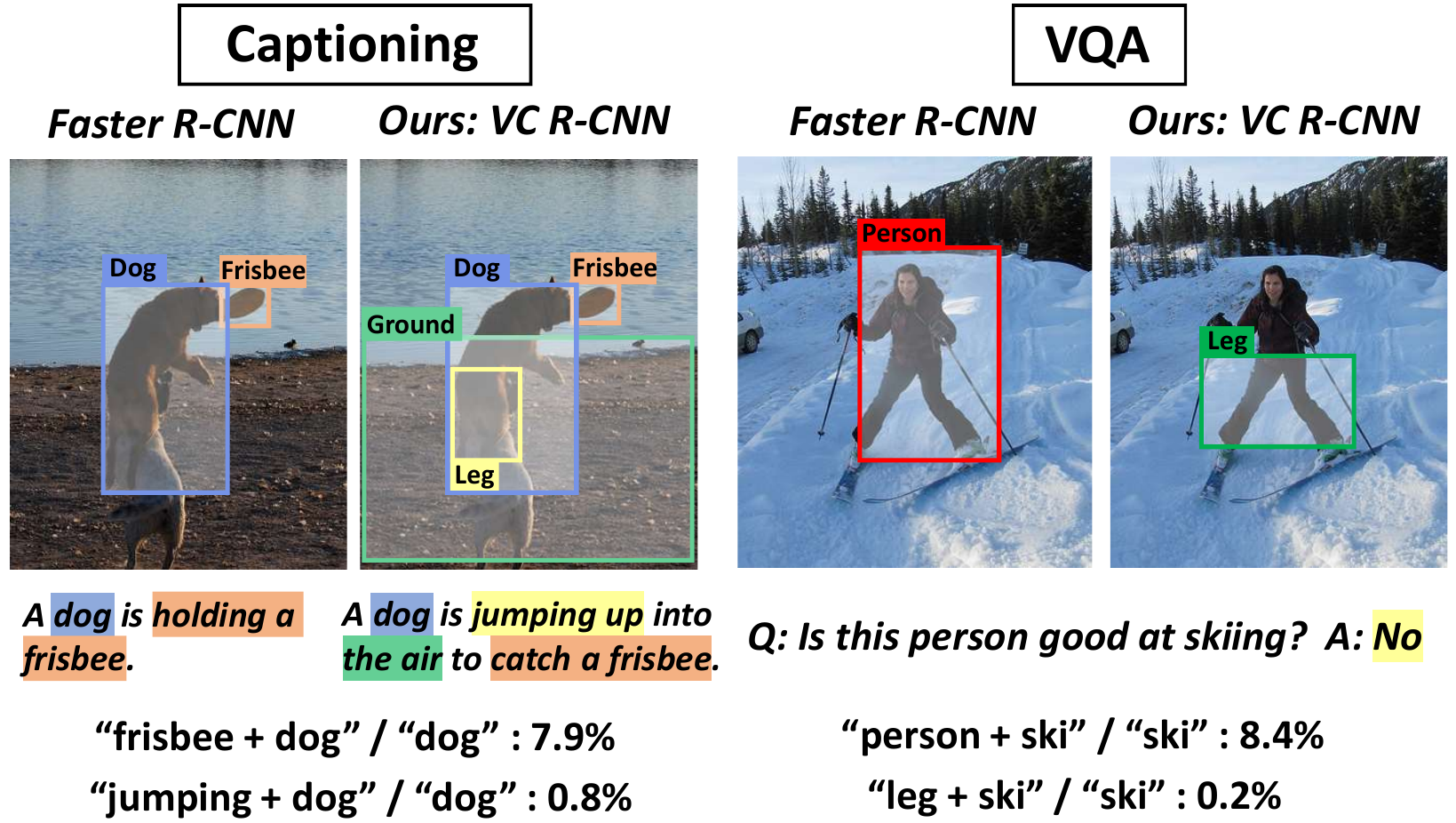}
\end{center}
  \caption{Examples of ``cognitive errors'' in image captioning and VQA due to the dataset bias. The ratio ./. denotes the co-occurrence\% in ground-truth text (captioning: captions, VQA: questions). By comparing with the Faster R-CNN~\cite{ren2015faster} based features~\cite{anderson2018bottom}, our VC R-CNN features can correct the errors, \eg, more accurate visual relationships and visual attentions, by being more commonsense awareness.}
\label{fig:figure1}
\vspace{-0.5cm}
\end{figure}

It is not hard to spot the ``cognitive errors'' committed by machines due to the lack of common sense. As shown in Figure~\ref{fig:figure1}, by using only the visual features, \eg, the prevailing Faster R-CNN~\cite{ren2015faster} based Up-Down~\cite{anderson2018bottom}, machine usually fails to describe the exact visual relationships (the captioning example), or, even if the prediction is correct, the underlying visual attention is not reasonable (the VQA example). Previous works blame this for dataset bias without further justification~\cite{hendricks2018women,manjunatha2019explicit,ramakrishnan2018overcoming,cadene2019rubi}, \eg, the large concept co-occurrence gap in Figure~\ref{fig:figure1}; but here we take a closer look at it by appreciating the difference between the ``visual'' and ``commonsense'' features. As the ``visual'' only tells ``what''/``where'' about \texttt{person} or \texttt{leg} \emph{per se}, it is just a more descriptive symbol than its correspondent English word; when there is bias, \eg, there are more \texttt{person} than \texttt{leg} regions co-occur with the word ``ski'', the visual attention is thus more likely to focus on the \texttt{person} region. On the other hand, if we could use the ``commonsense'' features, the action of ``ski'' can focuses on the \texttt{leg} region because of the common sense: we ski with legs.

We are certainly not the first to believe that visual features should include more commonsense knowledge, rather than just visual appearances. There is a trend in our community towards \emph{weakly-supervised} learning features from large-scale vision-language corpus~\cite{lu2019vilbert,sun2019videobert,tan2019lxmert}. However, despite the major challenge in trading off between annotation cost and noisy multimodal pairs, common sense is not always recorded in text due to the reporting bias~\cite{vedantam2015learning,lin2015don}, \eg, most may say ``people walking on road'' but few will point out ``people walking with legs''. In fact, we humans naturally learn common sense in an \emph{unsupervised fashion} by exploring the physical world, and we wish that machines can also imitate in this way.

A successful example is the unsupervised learning of word vectors in our sister NLP community~\cite{mikolov2013distributed, devlin2018bert, peters2018deep}: a word representation $X$ is learned by predicting its contextual word $Y$, \ie, $P(Y|X)$ in a neighborhood window. However, its counterpart in our own community, such as learning by predicting surrounding objects or parts~\cite{doersch2015unsupervised,malisiewicz2009beyond}, is far from effective in down-stream tasks. The reason is that the commonsense knowledge, in the form of language sentences, has already been recorded in discourse; in contrast, once an image has been taken, the explicit knowledge why objects are contextualized will never be observed, so the true common sense that \textbf{causes} the existence of objects $X$ and $Y$ might be \textbf{confounded} by the spurious \emph{observational bias}, \eg, if \texttt{keyboard} and \texttt{mouse} are more often observed with \texttt{table} than any other objects, the underlying common sense that \texttt{keyboard} and \texttt{mouse} are parts of \texttt{computer} will be wrongly attributed to \texttt{table}. 

\begin{figure}[t]
\begin{center}
\includegraphics[width=0.40\textwidth]{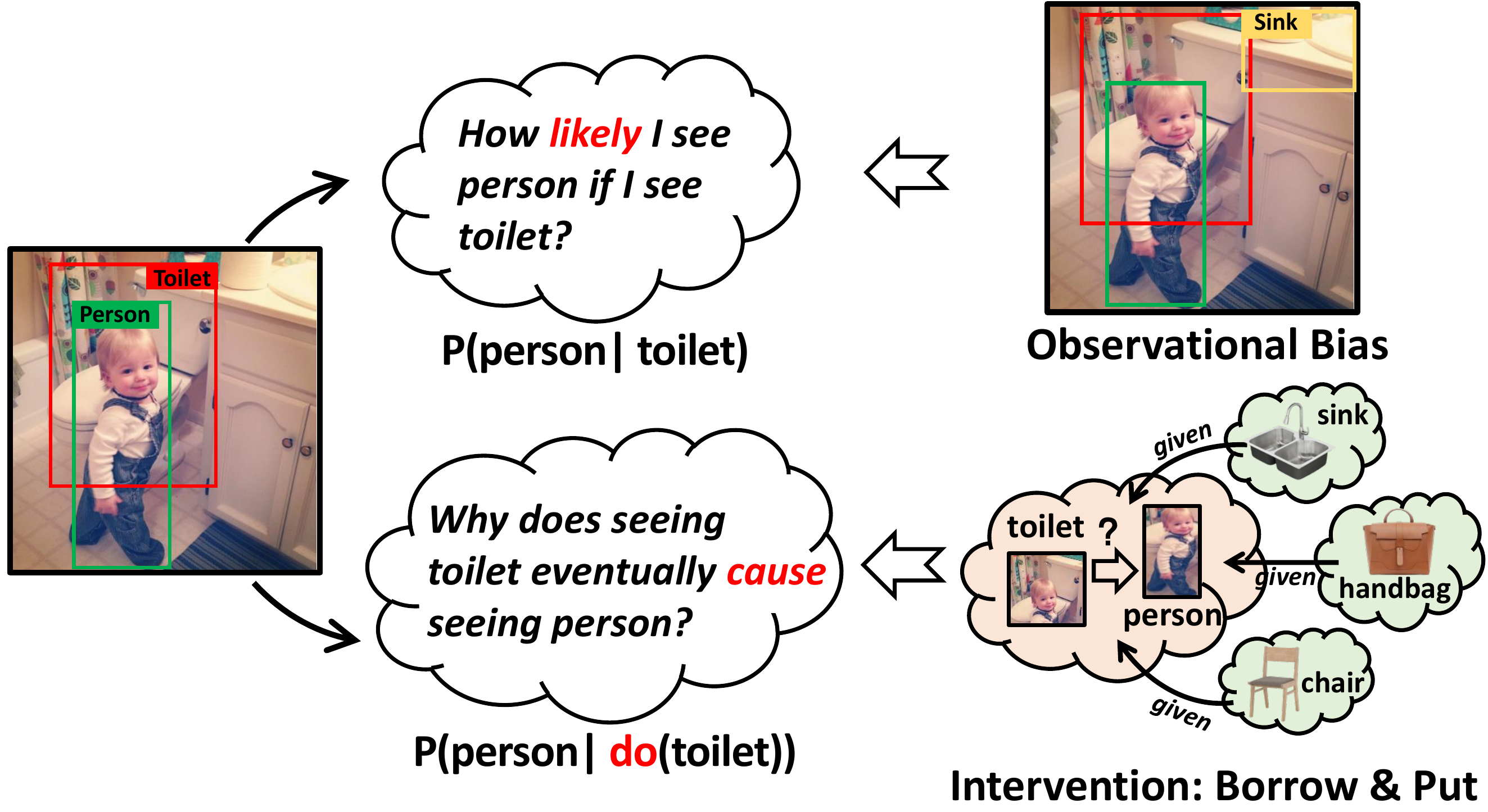}
\end{center}
\vspace{-0.2cm}
  \caption{The illustration of why $P(Y|do(X))$  learns common sense while $P(Y|X)$ does not. 
  Thanks to intervention, $P(Y|do(X))$ can ``borrow'' objects from other images and ``put'' them into the local image, to perform further justifications if $X$ truly causes $Y$ regardless of the unobserved confounders, and thus alleviate the observational bias.
  }
\label{fig:figure2}
\vspace{-0.4cm}
\end{figure}

\begin{figure}[t]
\begin{center}
\includegraphics[width=0.43\textwidth]{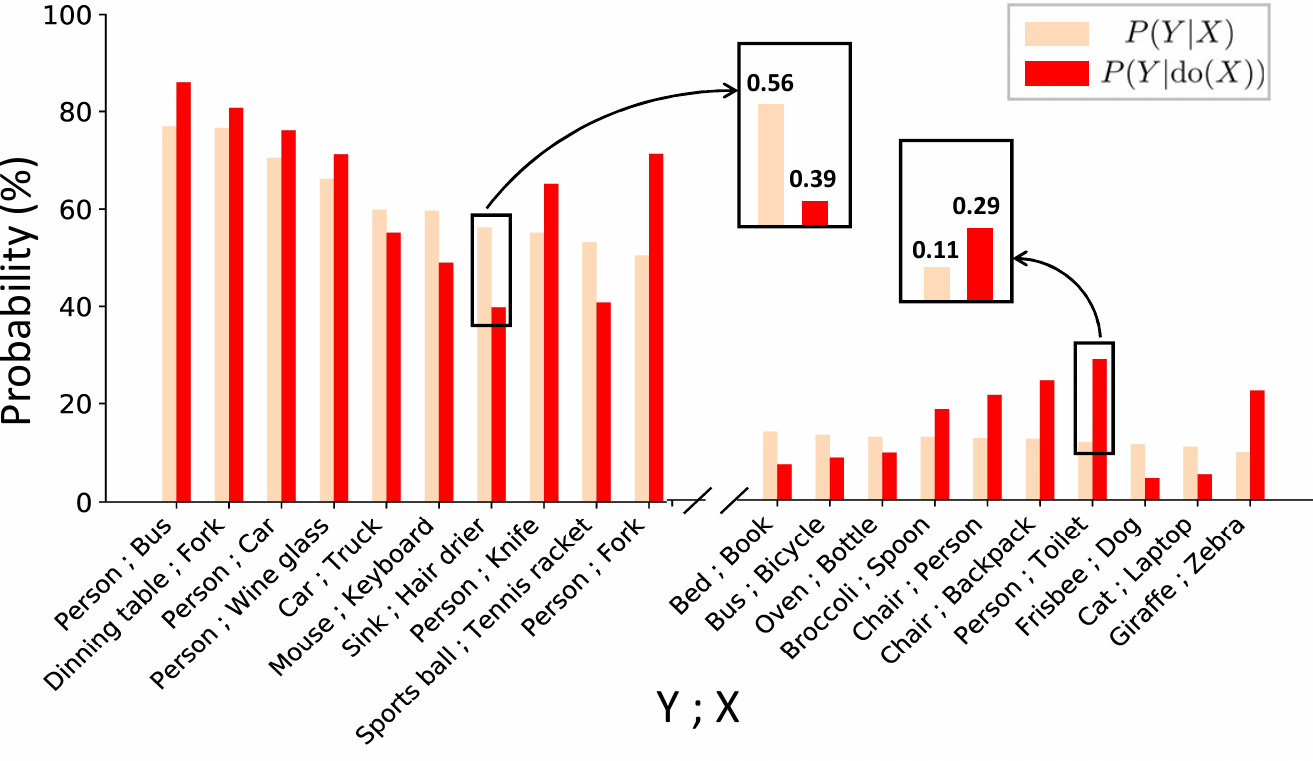}
\vspace{-0.3cm}
\end{center}
  \caption{The sensible difference between the likelihood before (\ie,$P(Y|X)$)  and after intervention (\ie, $P(Y|do(X))$) in MS-COCO. The object is represented by the 80 ground-truth class labels. Only 20 pairs are visualized to avoid clutter.}
\label{fig:figure3}
\vspace{-0.3cm}
\end{figure}

Intrigued, we perform a toy MS-COCO~\cite{lin2014microsoft} experiment with ground-truth object labels --- by using a mental apparatus, \emph{intervention}, that makes us human~\cite{pearl2018book} --- to screen out the existence of confounders and then eliminate their effect. We compare the difference between \emph{association} $P(Y|X)$ and  \emph{causal intervention} $P(Y|do(X))$~\cite{pearl2016causal}. Before we formally introduce \emph{do} in Section~\ref{sec:causal_intervention}, you can intuitively understand it as the following deliberate experiment illustrated in Figure~\ref{fig:figure2}: 1) ``borrow'' objects $Z$ from other images, 2) ``put'' them around $X$ and $Y$, then 3) test if $X$ still causes the existence of $Y$ given $Z$. The ``borrow'' and ``put'' is the spirit of intervention, implying that the chance of $Z$ is only dependent on us (probably subject to a prior), but independent on $X$ or $Y$. By doing so, as shown in Figure~\ref{fig:figure3}, $P(\texttt{sink}|do(\texttt{dryer}))$ is lower because the most common restroom context such as \texttt{towel} is forced to be seen as fair as others. Therefore, by using $P(Y|do(X))$ as the learning objective, the bias from the context will be alleviated. 

More intrigued, $P(\texttt{person}|do(\texttt{toilet}))$ is higher. Indeed, \texttt{person} and \texttt{toilet} co-occur rarely due to privacy. However, human's \emph{seeing} is fundamentally different from machine's because our instinct is to seek the \emph{causality} behind any association~\cite{pearl2018book} --- and here comes the common sense. As opposed to the passive observation $P(Y|X)$: ``How likely I see person if I see toilet'', we keep asking ``Why does seeing toilet eventually cause seeing person?'' by using $P(Y|do(X))$. Thanks to intervention, we can increase $P(Y|do(X))$ by ``borrowing'' non-local context that might not be even in this image, for the example in Figure~\ref{fig:figure2}, objects usable by \texttt{person} such as \texttt{chair} and \texttt{handbag} --- though less common in the restroom context --- will be still fairly ``borrowed'' and ``put'' in the image together with the common \texttt{sink}. 
We will revisit this example formally in Section~\ref{sec:causal_intervention}.

\begin{figure}[t]
\begin{center}
\includegraphics[width=0.45\textwidth]{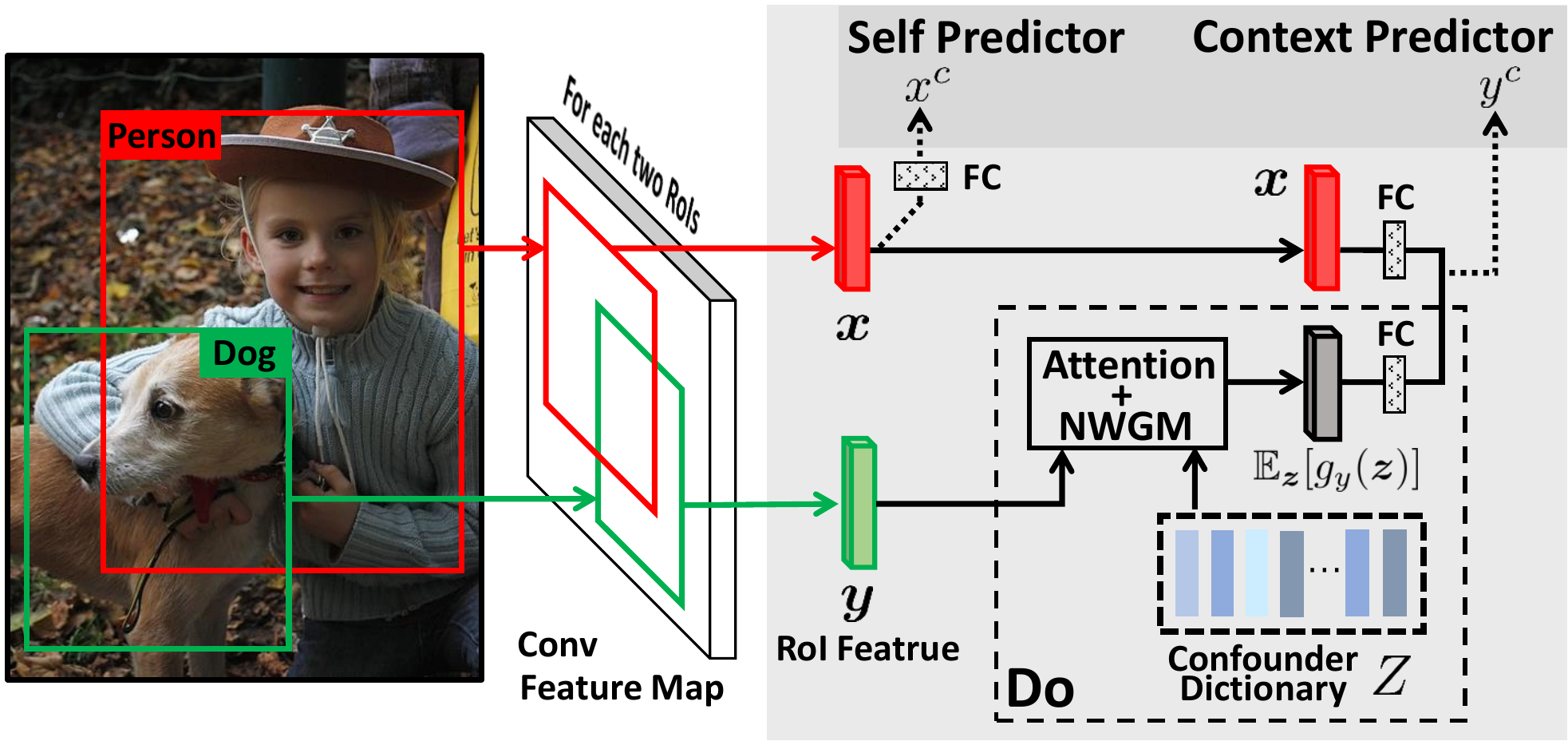}
\end{center}
   \caption{The overview of VC R-CNN. Any R-CNN backbone (\eg, Faster R-CNN~\cite{ren2015faster}) can be used to extract regions of interest (RoI) on the feature map. Each RoI is then fed into two sibling branches: a \textbf{Self Predictor} to predict its own class, \eg, $x^{c}$, and a \textbf{Context Predictor} to predict its context labels, \eg, $y^{c}$, with our \textbf{Do} calculus. The architecture is trained with a multi-task loss.}
\label{fig:framework}
\vspace{-0.1in}
\end{figure}

So far, we are ready to present our unsupervised region feature learning method: Visual Commonsense R-CNN (\textbf{VC R-CNN}), as illustrated in Figure~\ref{fig:framework}, which uses Region-based Convolutional Neural Network (R-CNN)~\cite{ren2015faster} as the visual backbone, and the causal intervention as the training objective. Besides its novel learning fashion, we also design a novel algorithm for the $do$-operation, which is an effective approximation for the imaginative intervention (cf. Section~\ref{sec:approximation}). 
The delivery of VC R-CNN is a region feature extractor for any region proposal, and thus it is fundamental and ready-to-use for many high-level vision tasks such as Image Captioning~\cite{vinyals2015show}, VQA~\cite{antol2015vqa}, and VCR~\cite{zellers2019recognition}. Through extensive experiments in Section~\ref{sec:exp}, VC R-CNN shows significant and consistent improvements over strong baselines --- the prevailing methods in each task. Unlike the recent ``Bert-like'' methods~\cite{lu2019vilbert, sun2019videobert} that require huge GPU computing resource for pre-training features and fine-tuning tasks, VC R-CNN is light and non-intrusive. By ``light'', we mean that it is just as fast and memory-efficient as Faster R-CNN~\cite{ren2015faster}; by ``non-intrusive'', we mean that re-writing the task network is not needed, all you need is \texttt{\color{blue}{numpy.concatenate}} and then ready to roll.

We apologize humbly to disclaim that VC R-CNN provides a philosophically correct definition of ``visual common sense''. We only attempt to step towards a \textbf{computational} definition in two intuitive folds: 1) common: unsupervised learning from the observed objects, and 2) sense-making: pursuing the causalities hidden in the observed objects. VC R-CNN not only re-thinks the conventional likelihood-based learning in our CV community, but also provides a promising direction --- causal inference~\cite{pearl2018book} --- via practical experiments.

\section{Related Work}

\noindent\textbf{Multimodal Feature Learning.}
With the recent success of pre-training language models (LM)~\cite{devlin2018bert, dai-etal-2019-transformer, peters2018deep} in NLP, several approaches~\cite{lu2019vilbert,sun2019videobert, tan2019lxmert,chen2019uniter} seek weakly-supervised learning from large, unlabelled multi-modal data to encode visual-semantic knowledge.
However, all these methods suffer from the reporting bias~\cite{vedantam2015learning,lin2015don} of language and the great memory cost for downstream fine-tuning. 
In contrast, our VC R-CNN is unsupervised learning only from images and the learned feature can be simply concatenated to the original representations.

\noindent\textbf{Un-/Self-supervised Visual Feature Learning}~\cite{domke2008killed,theis2015generative,malisiewicz2009beyond, kolesnikov2019revisiting, zhai2019s4l}. They aim to learn visual features through an elaborated proxy task such as denoising autoencoders~\cite{bengio2014deep,vincent2008extracting}, context \& rotation prediction~\cite{doersch2015unsupervised, gidaris2018unsupervised} and data augmentation~\cite{lee2019rethinking}. The context prediction is learned from correlation while image rotation and augmentation can be regarded as applying the random controlled trial~\cite{pearl2018book}, which is active and non-observational (physical); by contrast, our VC R-CNN learns from the observational causal inference that is passive and observational (imaginative).

\noindent\textbf{Visual Common Sense.}
Previous methods mainly fall into two folds: 1) learning from images with commonsense knowledge bases~\cite{vedantam2015learning, yatskar2016stating, sadeghi2015viske, su2018learning,wu2016ask,zhu2014reasoning} and 2) learning actions from videos~\cite{goyal2017something}.
However, the first one limits the common sense to the human-annotated knowledge, while the latter is essentially, again, learning from correlation.

\noindent\textbf{Causality in Vision.}
There has been a growing amount of efforts in marrying complementary strengths of deep learning and causal reasoning~\cite{pearl2016causal, pearl2014interpretation} and have been explored in several contexts, including image classification~\cite{chalupka2014visual,lopez2017discovering}, reinforcement learning~\cite{nair2019causal, dasgupta2019causal,bengio2019meta} and adversarial learning~\cite{kocaoglu2017causalgan,kalainathan2018sam}. Lately, we are aware of some contemporary works on visual causality such as visual dialog~\cite{qi2019two}, image captioning~\cite{yang2020deconfounded} and scene graph generation~\cite{tang2020unbiased}. Different from their task-specific causal inference, VC R-CNN offers a generic feature extractor.


\section{Sense-making by Intervention}
We detail the core technical contribution in VC R-CNN: causal intervention and its implementation.

\subsection{Causal Intervention}
\label{sec:causal_intervention}
\begin{figure}[htbp]
\vspace{-0.4em}
\begin{minipage}[c]{0.55\linewidth}
\includegraphics[width=1\linewidth]{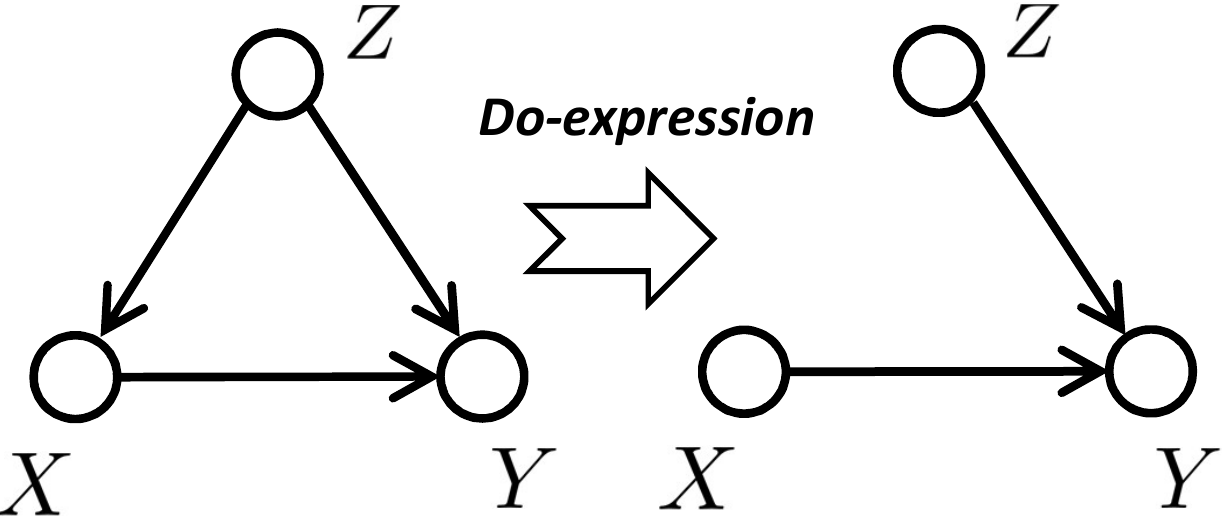}
\end{minipage}\hfill
\begin{minipage}[c]{0.4\linewidth}
\caption{\footnotesize The causal intervention $P(Y|do(X))$. Nodes denote variables and arrows denote the direct causal effects.}
\label{fig:do}
\end{minipage}\vspace{-3mm}
\end{figure}

\begin{figure}[t]
\centering
\includegraphics[width=0.43\textwidth]{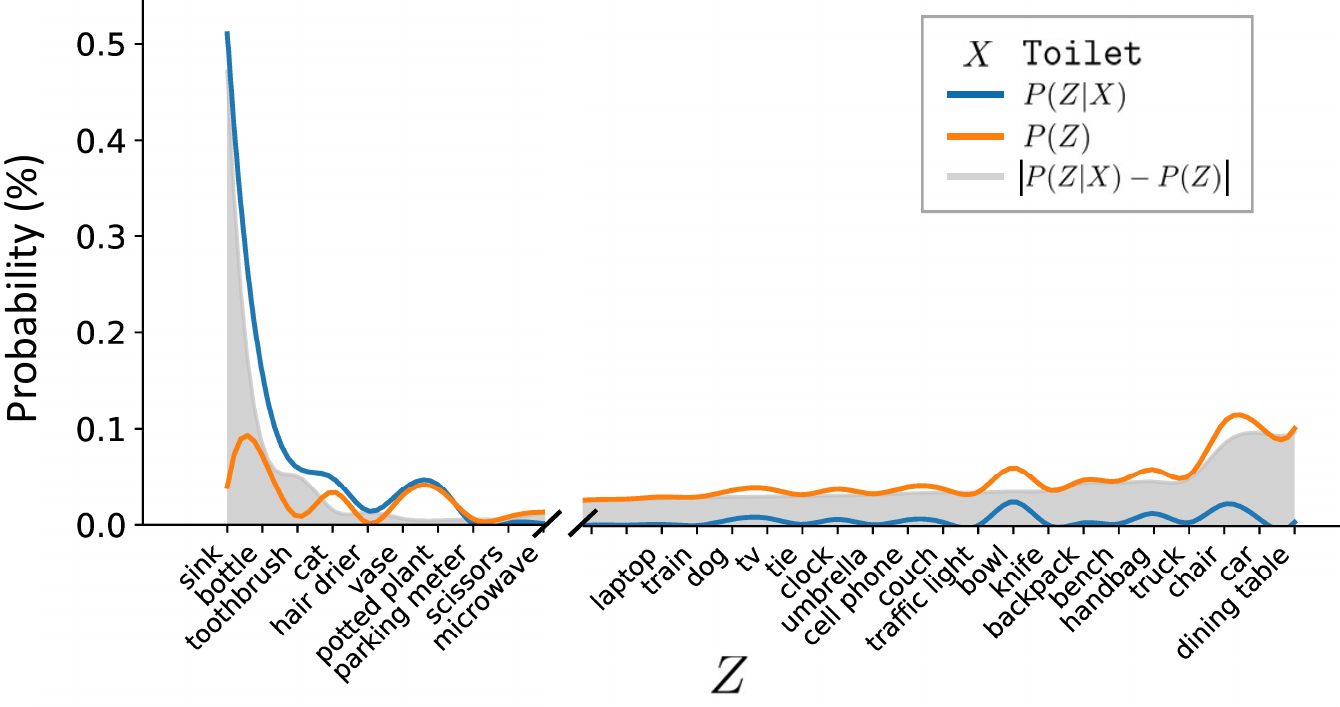}
\caption{A case study of the differences between $P(z|\texttt{Toilet}$ and $P(z)$ from MS-COCO ground-truth object labels. Only 29 labels of $Z$ are shown to avoid clutter.}
\label{fig:pzpzx}
\vspace{-0.06in}
\end{figure}

As shown in Figure~\ref{fig:do} (left), our visual world exists many confounders $z \in Z$ that affects (or causes) either $X$ or $Y$, leading to spurious correlations by only learning from the likelihood $P(Y|X)$. To see this, by using Bayes rule:
\begin{equation}
    P(Y | X) = \sum\nolimits_{z}P\left(Y| X, z\right)\underline{P\left(z|X\right)},
\label{eq:do_expression2}
\end{equation}
where the confounder $Z$ introduces the observational bias via $P(z|X)$. For example, as recorded in Figure~\ref{fig:pzpzx}, when $P$($z$\texttt{=sink}$|$$X$\texttt{=toilet}) is large while $P$($z$\texttt{=chair}$|$$X$\texttt{=toilet}) is small, most of the likelihood sum in Eq.~\eqref{eq:do_expression2} will be credited to $P$($Y$\texttt{=person}$|$$X$\texttt{=toilet},$z$\texttt{=sink}), other than $P$($Y$\texttt{=person}$|$$X$\texttt{=toilet},$z$\texttt{=chair}), so, the prediction from \texttt{toilet} to \texttt{person} will be eventually focused on \texttt{sink} rather than \texttt{toilet} itself, \eg, the learned features of a region \texttt{toilet} are merely its surrounding sink-like features.

As illustrated in Figure~\ref{fig:do} (right), if we intervene $X$, \eg, $do$($X$=\texttt{toilet}), the causal link between $Z$ and $X$ is cut-off. By applying the Bayes rule on the new graph, we have:
\vspace{-0.2cm}
\begin{equation}
    P(Y| do(X)) = \sum\nolimits_{z}P\left(Y | X, z\right)\underline{P(z)}.
\label{eq:do_expression1}
\vspace{-0.1cm}
\end{equation}
Compared to Eq.~\eqref{eq:do_expression2}, $z$ is no longer affected by $X$, and thus the intervention deliberately forces $X$ to incorporate every $z$ fairly, subject to its prior $P(z)$, into the prediction of $Y$. Figure~\ref{fig:pzpzx} shows the gap between the prior $P(z)$ and $P(z|\texttt{toilet})$, $z\in Z$ is the set of MS-COCO labels. We can use this figure to clearly explain the two interesting key results by performing intervention. Please note that $P(Y|X,z)$ remains the same in both Eq.~\eqref{eq:do_expression2} and Eq.~\eqref{eq:do_expression1}, 

Please recall Figure~\ref{fig:figure3} for the sensible difference between $P(Y|X)$ and $P(Y|do(X))$. First, $P$(\texttt{person}$|$\textit{do}(\texttt{toilet}))$>$$P$(\texttt{person}$|$\texttt{toilet}) is probably because the number of classes $z$ such that $P$($z|$\texttt{toilet})$>$$P(z)$ is smaller than those such that of $P$($z|$\texttt{toilet}) $<$ $P(z)$, \ie, the left grey area is smaller than the right grey area in Figure~\ref{fig:pzpzx}, making Eq.~\eqref{eq:do_expression2} smaller than Eq.~\eqref{eq:do_expression1}. Second, we can see that $z$ making $P(z)<P(z|X)$ is mainly from the common restroom context such as \texttt{sink}, \texttt{bottle}, and \texttt{toothbrush}. Therefore, by using intervention $P(Y|do(X))$ as the feature learning objective, we can adjust between ``common'' and  ``sense-making'', thus alleviate the observational bias.

\begin{figure}[t]
  \centering
    \subfloat[Object features learned by correlation $P(Y|X)$ and intervention $P(Y|do(X))$ (our VC R-CNN).]{\label{Fig:R1}\includegraphics[width=0.43\textwidth]{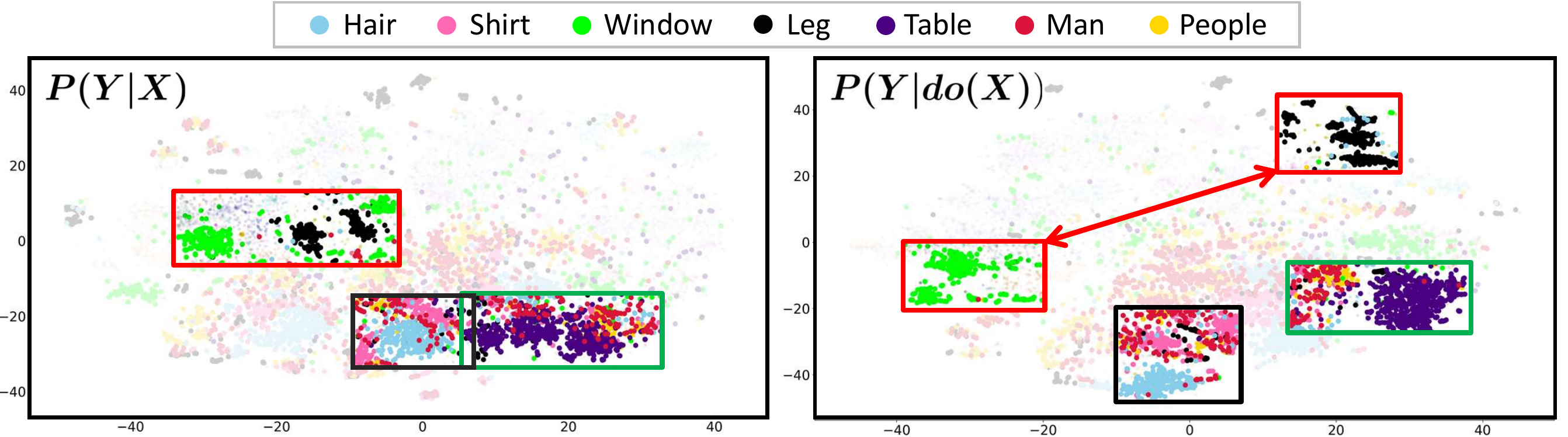}} \\ \vspace{-0.1in}
    \subfloat[Object features of Up-Down features and our VC R-CNN.]{\label{Fig:R2}\includegraphics[width=0.43\textwidth]{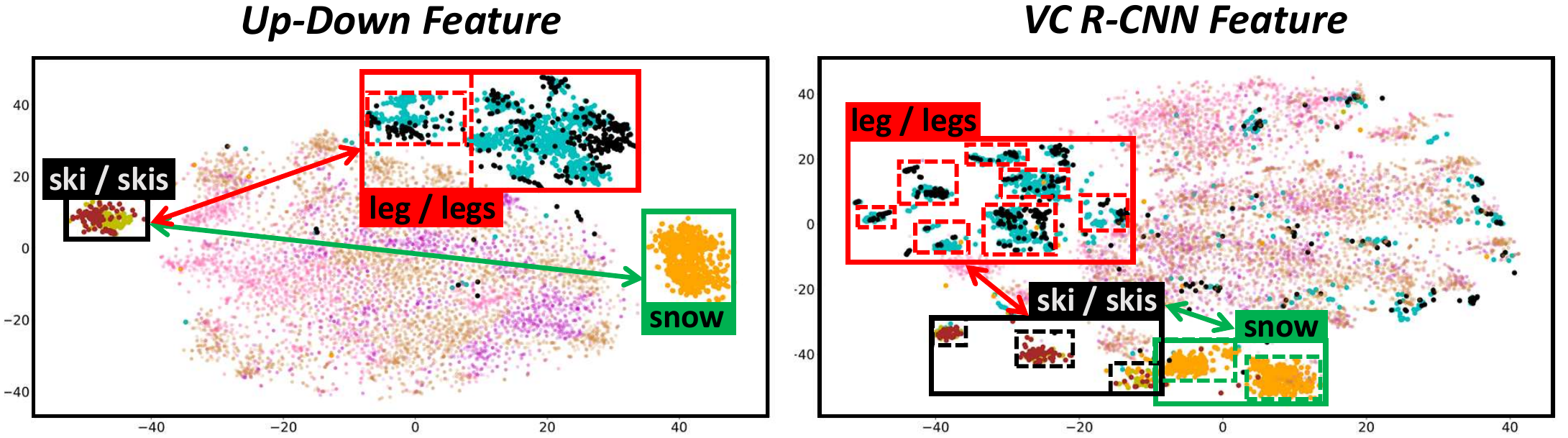}} \\
  \caption{The t-SNE visualization~\cite{maaten2008visualizing} of object features trained on MS-COCO with Up-Down~\cite{anderson2018bottom} provided Faster R-CNN labels. Features out of the label legend are faded out to avoid clutter.}
    \label{fig:tsne}
    \vspace{-0.1in}
\end{figure}

Figure~\ref{fig:tsne}\subref{Fig:R1} visualizes the features extracted from MS-COCO images by using the proposed VC R-CNN. Promisingly, compared to $P(Y|X)$ (left), $P(Y|do(X))$ (right) successfully discovers some sensible common sense. For example, before intervention, \texttt{window} and \texttt{leg} features in red box are close due to the street view observational bias, \eg, people walking on street with window buildings; after intervention, they are clearly separated. Interestingly, VC R-CNN \texttt{leg} features are closer to \texttt{head} while \texttt{window} features are closer to \texttt{wall}.  
Furthermore, Figure~\ref{fig:tsne}\subref{Fig:R2} shows the features of \texttt{ski}, \texttt{snow} and \texttt{leg} on same MS-COCO images via Up-Down (left) and our VC R-CNN (right).
We can see the \texttt{ski} feature of our VC R-CNN is reasonably closer to \texttt{leg} and \texttt{snow} than Up-Down.
Interestingly, VC R-CNN merges into sub-clusters (dashed boxes), implying that the common sense is actually multi-facet and varies from context to context.

\noindent\textbf{$\bm{X}$ $\rightarrow$ $\bm{Y}$ or $\bm{Y}$ $\rightarrow$ $\bm{X}$?}
We want to further clarify that both two causal directions between $X$ and $Y$ can be meaningful and indispensable with do calculus. For $X$ $\rightarrow$ $Y$, we want to learn the visual commonsense about $X$ (\egs, \texttt{toilet}) that causes the existence of $Y$ (\egs, \texttt{person}), and vice versa.\\
\noindent\textbf{Only objects are confounders?} No, some confounders are unobserved and beyond objects in visual commonsense learning, \eg, color, attributes, and the nuanced scene contexts induced by them; however, in unsupervised learning, we can only exploit the objects. Fortunately, this is reasonable: 1) we can consider the objects as the partially observed children of the unobserved confounder~\cite{d2019multi}; 2) we propose the implementation below to approximate the contexts, \eg, in Figure~\ref{fig:attn}, \texttt{Stop sign} may be the child of the confounder ``transportation'', and \texttt{Toaster} and \texttt{Refrigerator} may contribute to ``kitchen''.

\subsection{The Proposed Implementation}
\label{sec:approximation}
To implement the theoretical and imaginative intervention in Eq.~\eqref{eq:do_expression1}, we propose the proxy task of predicting the local context labels of $Y$'s RoI. For the confounder set $Z$, since we can hardly collect all confounders in real world, we approximate it to a fixed confounder dictionary $\bm{Z}=[\bm{z}_1,...,\bm{z}_N]$ in the shape of $N \times d$ matrix for practical use, where $N$ is the category size in dataset (\egs, 80 in MS-COCO) and $d$ is the feature dimension of RoI. Each entry $\bm{z}_i$ is the averaged RoI feature of the $i$-th category samples in dataset. The feature is pre-trained by Faster R-CNN.

Specifically, given $X$'s RoI feature $\bm{x}$ and its contextual $Y$'s RoI whose class label is $y^c$, Eq.~(\ref{eq:do_expression1}) can be implemented as $\sum_{z}P\left(y^{c} | \bm{x}, \bm{z}\right)P\left(\bm{z}\right)$.
The last layer of the network for label prediction is the Softmax layer: $P\left(y^{c} | \bm{x}, \bm{z}\right) = \textit{Softmax}(f_{y}(\bm{x},\bm{z}))$, where $f_{y}(\cdot)$ calculates the logits for $N$ categories, and the subscript $y$ denotes that $f(\cdot)$ is parameterized by $Y$'s RoI feature $\bm{y}$, motivated by the intuition that the prediction for $y^c$ should be characterized by $Y$. In summary, the implementation is defined as:
\vspace{-0.08cm}
\begin{equation}
    P\left(Y | do(X)\right)
    := \mathbb{E}_{\bm{z}}[\textit{Softmax}(f_{y}(\bm{x},\bm{z}))].
\label{eq:do_expression3}
\vspace{-0.08cm}
\end{equation}
Note that $\mathbb{E}_{\bm{z}}$ requires expensive sampling.

\noindent\textbf{Normalized Weighted Geometric Mean (NWGM).}
We apply NWGM~\cite{xu2015show} to approximate the above expectation. In a nutshell, NWGM\footnote{The detailed derivation about NWGM can be found in the Supp..} effeciently moves the outer expectation into the Softmax as:
\vspace{-0.08cm}
\begin{equation}
    \mathbb{E}_{\bm{z}}[\textit{Softmax}(f_{y}(\bm{x},\bm{z}))]\!
    \overset{\text{NWGM}}{\approx}\! \textit{Softmax}(\mathbb{E}_{\bm{z}}[f_{y}(\bm{x},\bm{z})]).
\label{eq:nwgm}
\vspace{-0.08cm}
\end{equation}
In this paper, we use the linear model $f_{y}(\bm{x}, \bm{z}) = \bm{W}_1\bm{x} + \bm{W}_2\cdot g_{y}(\bm{z})$, where $\bm{W}_{1}, \bm{W}_{2}\in \mathbb{R}^{N \times d}$ denote the fully connected layer.
Then the Eq.~(\ref{eq:nwgm}) can be derived as:
\vspace{-0.08cm}
\begin{equation}
\mathbb{E}_{\bm{z}}[f_{y}(\bm{x},\bm{z})] = \bm{W}_{1}\bm{x} + \bm{W}_{2}\cdot \mathbb{E}_{\bm{z}}[g_{y}(\bm{z})].
\label{eq:ez}
\vspace{-0.08cm}
\end{equation}
Note that the above approximation is reasonable, because the effect on $Y$ comes from both $X$ and confounder $Z$ (cf. the right Figure~\ref{fig:do}).

\begin{figure}[t]
\begin{center}
\includegraphics[width=0.47\textwidth]{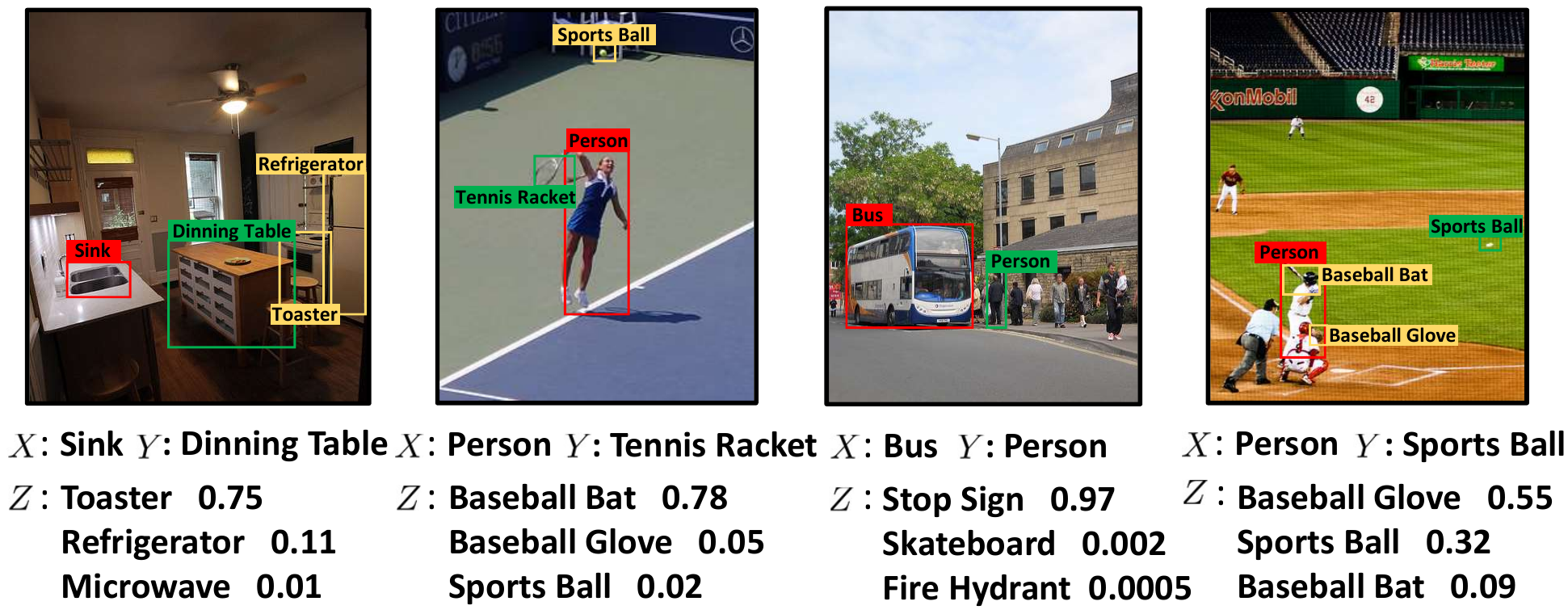}
\end{center}
\vspace{-0.2cm}
   \caption{The visualizations of the top 3 confounders given RoI feature $\bm{x}$ (red box) and $\bm{y}$ (green box), while numbers denote the attention weight. We can see that our model can recognize reasonable confounders $\bm{z}$, \eg, the common context (yellow boxes).} 
\label{fig:attn}
\vspace{-0.1in}
\end{figure}

\noindent\textbf{Computing $\mathbb{E}_{\bm{z}}[g_{y}(\bm{z})]$.}
Specifically, given $\bm{y}$ and confounder dictionary $\bm{Z}$, 
we firstly calculate the attention vector $\bm{a} = \textit{Softmax}(\bm{q}^{T}\bm{K} / \sqrt{\sigma})$, followed by the broadcasting operation to get matrix $\bm{A}=[\bm{a};...;\bm{a}]$ of the same shape as $\bm{Z}$, where $[;]$ denotes broadcasting along the row.
$\bm{q}=\bm{W}_{3}\bm{y}$, $\bm{K}=\bm{W}_{4} \bm{Z}^{T}$.
$\bm{W}_{3}$, $\bm{W}_{4}$ map each vector to the common subspace and
$\sigma$ denotes the first dimension of $\bm{W}_{3}$, $\bm{W}_{4}$ as a constant scaling factor.
Then we can have $\mathbb{E}_{\bm{z}}[g_{y}(\bm{z})] = \sum\nolimits_{z}[\bm{A} \odot \bm{Z}] P(\bm{z})$, where $P(\bm{z})$ denotes the prior statistic probability and $\odot$ is the element-wise product.
Figure~\ref{fig:attn} visualizes the top 3 confounders ranked by the soft attention weights. Note that they are the cancer in learning ``sense-making'' features from $P(Y|X)$.

\noindent\textbf{Neural Causation Coefficient (NCC)}.
Due to the fact that the causality from the confounders as the category averaged features are not yet verified, that is,  $\bm{Z}$ may contain colliders (or v-structure)~\cite{pearl2016causal} causing spurious correlations when intervention. To this end, we apply NCC~\cite{lopez2017discovering} to remove possible colliders from $\bm{Z}$.
Given $\bm{x}$ and $\bm{z}$, $NCC\left(\bm{x} \rightarrow \bm{z}\right)$  outputs the relative causality intensity from $\bm{x}$ to $\bm{z}$. Then we discard the training samples with strong collider causal intensities above a threshold.

\section{VC R-CNN}

\noindent\textbf{Architecture.}
Figure~\ref{fig:framework} illustrates the VC R-CNN architecture. 
VC R-CNN takes an image as input and generates feature map from a CNN backbone (\eg, ResNet101~\cite{he2016deep}). 
Then, unlike Faster R-CNN~\cite{ren2015faster}, we discard the Region Proposal Network (RPN).
The ground-truth bounding boxes are directly utilized to extract the object level representation with the RoIAlign layer. Finally, each two RoI features $\bm{x}$ and $\bm{y}$ eventually branch into two sibling predictors: Self Predictor with a fully connected layer to estimate each object class, while Context Predictor with the approximated \emph{do}-calculus in Eq.~\eqref{eq:do_expression3} to predict the context label.

\noindent\textbf{Training Objectives.}
The Self-Predictor outputs a discrete probability distribution $p=(p[1],...,p[N])$ over $N$ categories (note that we do not have the ``background'' class).
The loss can be defined as $L_{self}(p, x^{c}) = -log(p[x^c])$, where $x^{c}$ is the ground-truth class of RoI $X$. The Context Predictor loss $L_{cxt}$ is defined for each two RoI feature vectors. Considering $X$ as the center object while $Y_i$ is one of the $K$ context objects with ground-truth label $y_i^{c}$,
the loss is $L_{cxt}(p_i, y_i^{c}) = -log(p_{i}[y_i^c])$, where $p_{i}$ is calculated by $p_i = P(Y_i|do(X))$ in Eq.~\eqref{eq:do_expression3} and  $p_i=(p_i[1],...,p_i[N])$ is the probability over $N$ categories.
Finally, the overall mulit-task loss for each RoI $X$ is:
\begin{equation}
    L\left(X\right)
    = L_{self}(p, x^{c}) + \frac{1}{K}\sum\nolimits_{i}L_{cxt}(p_i, y_i^{c}).
\label{eq:loss}
\end{equation}

\noindent\textbf{Feature Extractor.}
We consider VC R-CNN as a visual commonsense feature extractor for any region proposal. Then the extracted features are directly concatenated to the original visual feature utilized in any downstream tasks. It is worth noting that we do NOT recommend early concatenations for some models that contain a self-attention architecture such as AoANet~\cite{huang2019attention}. The reasons are two-fold. First, as the computation of these models are expensive, early concatenation significantly slows down the training. Second, which is more crucial, the self-attention essentially and implicitly applies $P(Y|X)$, which contradicts to causal intervention. We will detail this finding in Section~\ref{sec:results}.

\section{Experiments}
\label{sec:exp}

\subsection{Datasets}
We used the two following datasets for unsupervised learning VC R-CNN.

\noindent\textbf{MS-COCO Detection}~\cite{lin2014microsoft}. It is a popular benchmark dataset for classification, detection and segmentation in our community. It contains 82,783, 40,504 and 40,775 images for training, validation and testing respectively with 80 annotated classes.
Since there are 5K images from downstream image captioning task which can be also found in MS-COCO validation split, we removed those in training. Moreover, recall that our VC R-CNN relies on the context prediction task, thus, we discarded images with only one annotated bounding box.

\noindent\textbf{Open Images~\cite{kuznetsova2018open}.}
We also used a much larger dataset called Open Images, a huge collection containing 16M bounding boxes across 1.9M images, making it the largest object detection dataset.
We chose images with more than three annotations from the official training set, results in about 1.07 million images consisting of 500 classes.

\begin{table}[t]
\centering
\scalebox{0.73}{
\begin{tabular}{p{0.8cm}p{1.5cm}p{0.5cm}<{\centering}p{0.5cm}<{\centering}p{0.5cm}<{\centering}p{0.6cm}<{\centering}p{0.5cm}<{\centering}p{0.5cm}<{\centering}p{0.5cm}<{\centering}p{0.6cm}<{\centering}}
\hline\hline
\multirow{2}{*}{\large{Model}}  & \multicolumn{1}{c}{\multirow{2}{*}{\large{Feature}}}      & \multicolumn{4}{c}{\textbf{MS-COCO}}              &\multicolumn{4}{c}{\textbf{Open Images}}         \\ \cmidrule(lr){3-6}\cmidrule(lr){7-10}
& \multicolumn{1}{c}{}       & B4   & M    & R      & C    & B4   & M    & R       & C     \\ \hline
\multicolumn{1}{c}{\multirow{6}{*}{\rotatebox{90}{\large{Up-Down}}}} & Origin~\cite{anderson2018bottom} & 36.3 & 27.7 & 56.9  & 120.1  & 36.3 & 27.7 & 56.9  & 120.1 \\
                     & Obj      & 36.7 & 27.8 & 57.5  & 122.3  & 36.7 & 27.8 & 57.5  & 122.3 \\
                      & Only VC           &34.5  &27.1   &56.5     &115.2      &35.1    &27.2      &56.6    & 115.7          \\
                       & +Det  & 37.5 & 28.0   & 58.3  & 125.9  & 37.4 & 27.9 & 58.2  & 125.7 \\
                     & +Cor  & 38.1 & 28.3 & 58.5    & 127.5  & 38.3 & 28.4 & 58.8  & 127.4 \\
                    & \cellcolor{mygray}+VC    &\cellcolor{mygray}39.5 & \cellcolor{mygray}29.0 & \cellcolor{mygray}59.0  & \cellcolor{mygray}130.5     & \cellcolor{mygray}39.1 & \cellcolor{mygray}28.8 & \cellcolor{mygray}59.0   & \cellcolor{mygray}130.0 \\ \hline
\multicolumn{1}{c}{\multirow{6}{*}{\rotatebox{90}{\large{AoANet$^{\dagger}$}}}}  & Origin\tablefootnote{Since we cannot achieve performances reported in original paper using the official code even with the help of author, here we show ours as the baseline. The original results can be found at the bottom row: SOTA.\label{ftnote}}~\cite{huang2019attention}  & 38.9 & 28.9 & 58.8  & 128.4 & 38.9 & 28.9 & 58.8  & 128.4 \\
                     & Obj     & 38.1 & 28.4 & 58.2  & 126.0   & 38.1 & 28.4 & 58.2  & 125.9 \\
                      & Only VC          &35.8     &27.6      & 56.8       &118.1               & 35.8     & 27.9     &56.7         & 118.5      \\
                     & +Det  & 38.8 & 28.8 & 58.7  & 128.0    & 38.7 & 28.6 & 58.7  & 127.7 \\
                     & +Cor & 38.8 & 28.9 & 58.7  & 128.6  & 38.9 & 28.8 & 58.7  & 128.2 \\
                    & \cellcolor{mygray}+VC    & \cellcolor{mygray}\textbf{39.5} & \cellcolor{mygray}\textbf{29.3} & \cellcolor{mygray}\textbf{59.3}  & \cellcolor{mygray}\textbf{131.6}  & \cellcolor{mygray}39.3 & \cellcolor{mygray}29.1   & \cellcolor{mygray}59.0   & \cellcolor{mygray}131.5 \\ \hline
SOTA                     & AoANet~\cite{huang2019attention}           & 38.9 & 29.2 & 58.2  & 129.8  & 38.9 & 29.2 & 58.2 & 129.8 \\ \hline \hline
\end{tabular}}
\caption{The image captioning performances of representative two models with ablative features on Karpathy split. The metrics: B4, M, R and C denote BLEU@4, METEOR, ROUGE-L and CIDEr-D respectively. The grey row highlight our features in each model. AoANet$^{\dagger}$ indicates the AoANet without the refine encoder. Note that the Origin and Obj share the same results in MS-COCO and Open Images since they does not contain our new trained features.}
\label{tab:caption}
\vspace{-0.3cm}
\end{table}

\subsection{Implementation Details}

We trained our VC R-CNN on 4 Nvidia 1080Ti GPUs with a total batch size of 8 images for 220K iterations (each mini-batch has 2 images per GPU). 
The learning rate was set to 0.0005 which was decreased by 10 at 160K and 200K iteration. ResNet-101 was set to the image feature extraction backbone. We used SGD as the optimizer with weight decay of 0.0001 and momentum of 0.9 following~\cite{ren2015faster}. To construct the confounder dictionary $\bm{Z}$, we first employed the pre-trained official ResNet-101 model on Faster R-CNN with ground-truth boxes as the input to extract the RoI features for each object. For training on Open Images, we first trained a vanilla Faster R-CNN model.
Then $\bm{Z}$ is built by making average on RoIs of the same class and is fixed during the whole training stage.

\subsection{Comparative Designs}
To evaluate the effectiveness of our VC R-CNN feature (VC), we present three representative vision-and-language downstream tasks in our experiment.
For each task, a \textbf{classic} model and a \textbf{state-of-the-art} model were both performed for comprehensive comparisons.
For each method, we used the following five ablative feature settings: 
1) \textbf{Obj}: the features based on Faster R-CNN, we adopted the popular used bottom-up feature~\cite{anderson2018bottom}; 2)  \textbf{Only VC}: pure VC features; 3) \textbf{+Det}: the features from training R-CNN with single self detection branch without Context Predictor. ``+'' denotes the extracted features are concatenated with the original feature, \egs, bottom-up feature; 4) \textbf{+Cor}: the features from training R-CNN by predicting all context labels (\ies, correlation) without the intervention; 5) \textbf{+VC}: our full feature with the proposed implemented intervention, concatenated to the original feature.
For fair comparisons, we retained all the settings and random seeds in the downstream task models.
Moreover, since some downstream models may have different settings in the original papers, we also quoted their results for clear comparison. For each downstream task, we detail the problem settings, dataset and evaluation metrics as below.

\begin{table}[t]
\centering
\scalebox{0.72}{
\begin{tabular}{lcccccccc}
\hline\hline
Model      & \multicolumn{2}{c}{BLEU-4} & \multicolumn{2}{c}{METEOR} & \multicolumn{2}{c}{ROUGE-L} & \multicolumn{2}{c}{CIDEr-D} \\ \hline
Metric     & c5           & c40         & c5           & c40         & c5           & c40          & c5           & c40          \\ \hline
Up-Down~\cite{anderson2018bottom}    & 36.9         & 68.5        & 27.6         & 36.7        & 57.1         & 72.4         & 117.9        & 120.5        \\
SGAE~\cite{yang2019auto}       & 37.8         & 68.7        & 28.1         & 37.0          & 58.2         & 73.1         & 122.7        & 125.5        \\
CNM~\cite{yang2019learning}        & 37.9         & 68.4        & 28.1         & 36.9        & 58.3         & 72.9         & 123.0        & 125.3        \\
AoANet~\cite{huang2019attention}     & 37.3         & 68.1        & 28.3         & 37.2        & 57.9         & 72.8         & 124.0        & 126.2        \\ \hline
Up-Down+VC & 37.8         & 69.1        & 28.5         & 37.6        & 58.2         & 73.3         & 124.1        & 126.2        \\
AoANet$^{\dagger}$+VC  & \textbf{38.4}         & \textbf{69.9}        & \textbf{28.8}         & \textbf{38.0}        & \textbf{58.6}         & \textbf{73.8}         & \textbf{125.5}        & \textbf{128.1}        \\ \hline\hline
\end{tabular}}
\caption{The performances of various single models on the online MS-COCO test server. Up-Down+VC and AoANet$^{\dagger}$+VC are the short for concatenated on~\cite{anderson2018bottom} in Up-Down and AoANet$^{\dagger}$.}
\label{tab:caption_test}
\end{table}

\begin{table}[]
\centering
\scalebox{0.75}{
\begin{tabular}{clccclcc}
\hline\hline
Model                   & Feature         & CHs  & Chi & Model                  & Feature         & CHs  & Chi \\ 
\hline
\multirow{4}{*}{\rotatebox{90}{Up-Down}} & Obj              & 12.8  &8.1  & \multirow{4}{*}{\rotatebox{90}{AoANet$^{\dagger}$}} & Obj              &12.6   &8.0  \\
                         & +Det  & 12.0   & 7.5 &                         & +Det  & 9.5  & 6.2 \\
                         & +Cor  & 11.2 & 7.1 &                         & +Cor  & 10.4 & 6.5 \\
                         & +VC  & \textbf{10.3} & \textbf{6.5} &                         & +VC  & \textbf{8.8}  & \textbf{5.5} \\ \hline\hline
\end{tabular}}
\caption{Hallucination analysis~\cite{rohrbach2018object} of various models on MS-COCO Karpathy test split to measure object hallucination for image captioning. The lower, the better.}
\label{tab:bias}
\vspace{-0.4cm}
\end{table}


\noindent\textbf{Image Captioning.}
Image captioning aims to generate textual description of an image.
We trained and evaluated on the most popular ``Karpathy'' split built on MS-COCO dataset, where 5K images for validation, 5K for testing, and the rest for training.
The sentences were tokenized and changed to lowercase.
Words appearing less than 5 times were removed and each caption was trimmed to a maximum of 16 words.
Five standard metrics were applied for evaluating the performances of the testing models: CIDEr-D~\cite{vedantam2015cider}, BLEU~\cite{papineni2002bleu}, METROT~\cite{banerjee2005meteor}, ROUGE~\cite{lin2004rouge} and SPICE~\cite{anderson2016spice}.

\noindent\textbf{Visual Question Answering (VQA).}
The VQA task requires answering natural language questions according to the images.
We evaluated the VQA model on VQA2.0~\cite{goyal2017making}. 
Compared with VQA1.0~\cite{antol2015vqa}, VQA2.0 has more question-image pairs for training (443,757) and validation (214,354), and all the question-answer pairs are balanced.
Before training, we performed standard text pre-processing.
Questions were trimed to a maximum of 14 words and candidate answer set was restricted to answers appearing more than 8 times.
The evaluation metrics consist of three pre-type accuracies (\ies, ``Yes/No'', ``Number'' and ``Other'').

\noindent\textbf{Visual Commonsense Reasoning (VCR).}
In VCR, given a challenging question about an image, machines need to present two sub-tasks: answer correctly (Q$\rightarrow$A) and provide a rationale justifying its answer (QA$\rightarrow$R).
The VCR dataset~\cite{zellers2019recognition} contains over 212K (training), 26K (validation) and 25K (testing) derived from 110K movie scenes.
The model was evaluated in terms of 4-choice accuracy and the random guess accuracy on each sub-task is 25\%.

\begin{table}[]
\centering
\scalebox{0.73}{
\begin{tabular}{p{0.8cm}lp{0.5cm}p{0.5cm}p{0.5cm}p{0.5cm}p{0.5cm}p{0.5cm}p{0.5cm}p{0.5cm}}
\hline\hline
\multirow{2}{*}{\large{Model}}   & \multicolumn{1}{c}{\multirow{2}{*}{\large{Feature}}} & \multicolumn{4}{c}{\textbf{MS-COCO}}     & \multicolumn{4}{c}{\textbf{Open Images}} \\ \cmidrule(lr){3-6}\cmidrule(lr){7-10}
                         & \multicolumn{1}{c}{}                         & Y/N & Num & Other & All  & Y/N  & Num & Other & All  \\ \hline
\multicolumn{1}{c}{\multirow{5}{*}{\rotatebox{90}{\large{Up-Down}}}} & Obj~\cite{anderson2018bottom}                                 & 80.3   & 42.8   & 55.8  & 63.2 & 80.3    & 42.8   & 55.8  & 63.2 \\
                         & Only VC                                     &77.8     &37.9   &51.6   &59.8  &77.9    &38.1    &51.1 & 59.9   \\
                         & +Det                             & 81.8   & 44.5   & 56.8  & 64.5 & 81.9    & 44.7   & 56.5  & 64.6 \\
                         & +Cor                             & 81.5   & 44.6   & 57.1  & 64.7 & 81.3    & 44.7   & 57.0    & 64.6 \\
                         & \cellcolor{mygray}+VC                              & \cellcolor{mygray}82.5   & \cellcolor{mygray}46.0   & \cellcolor{mygray}57.6  & \cellcolor{mygray}65.4 & \cellcolor{mygray}82.8    & \cellcolor{mygray}45.7   & \cellcolor{mygray}57.4  & \cellcolor{mygray}65.4 \\ \hline
\multicolumn{1}{c}{\multirow{5}{*}{\rotatebox{90}{\large{MCAN}}}}    & Obj~\cite{yu2019deep}                                 & 84.8   & \textbf{49.4}   & 58.4  & 67.1 & 84.8    & \textbf{49.4}   & 58.4  & 67.1 \\
                         & Only VC                             &80.8    &40.7    &48.9    &60.1  & 81.0   & 40.8  & 49.1    & 60.3 \\
                         & +Det                             & 84.8   & 49.2   & 58.8  & 67.2 &84.9     &49.3    &58.4  &67.2      \\
                         & +Cor                             & 85.0     & 49.2  & 58.9  & 67.4 & 85.1    & 49.1    & 58.6 &67.3      \\
                         & \cellcolor{mygray}+VC         
& \cellcolor{mygray}\textbf{85.2}   & \cellcolor{mygray}\textbf{49.4}   & \cellcolor{mygray}\textbf{59.1}  & \cellcolor{mygray}\textbf{67.7} & \cellcolor{mygray}85.1    &\cellcolor{mygray} 49.1   &\cellcolor{mygray} 58.9  &\cellcolor{mygray} 67.5 \\ \hline
SOTA & MCAN
& 84.8   & \textbf{49.4}   & 58.4  & 67.1 & 84.8    & \textbf{49.4}   & 58.4  & 67.1 \\ \hline\hline
\end{tabular}}
\caption{Accuracy (\%) of various ablative features on VQA2.0 validation set. Since the Obj achieves almost equal results with that in the original paper, here we just merge the two rows.}
\label{tab:vqa}
\end{table}

\begin{table}[]
\centering
\scalebox{0.73}{
\begin{tabular}{lccccc}
\hline\hline
\multicolumn{1}{c}{\multirow{2}{*}{\large{Model}}} & \multicolumn{4}{c}{test-dev}  & test-std \\ 
\cmidrule(lr){2-5}\cmidrule(lr){6-6}
\multicolumn{1}{c}{}                       & Y/N   & Num   & Other & All   & All      \\ \hline
Up-Down~\cite{anderson2018bottom}                                    & 81.82 & 44.21 & 56.05 & 65.32 & 65.67    \\
BAN~\cite{Kim2018}                                        & 85.46 & 50.66 & 60.50  & 69.66 & -        \\
DFAF~\cite{gao2019dynamic}                                       & 86.09 & 53.32 & 60.49 & 70.22 & 70.34    \\
MCAN~\cite{yu2019deep}                                       & 86.82 & 54.04 & 60.52 & 70.63 & 70.90     \\
\hline
UP-Down+VC                                 & 84.26 & 48.50  & 58.86 & 68.15 &68.45     \\
MCAN+VC                                    & \textbf{87.41} & \textbf{53.28} & \textbf{61.44} & \textbf{71.21} & \textbf{71.49}    \\ \hline\hline
\end{tabular}}
\caption{Single model accuracies (\%) on VQA2.0 test-dev and test set, where Up-Down+VC and MCAN+VC are the short for Object-VC R-CNN feature in Up-Down and MCAN.}
\label{tab:vqa_test}
\vspace{-0.5cm}
\end{table}


\begin{figure*}[!htb]
\begin{center}
\includegraphics[width=0.85\textwidth]{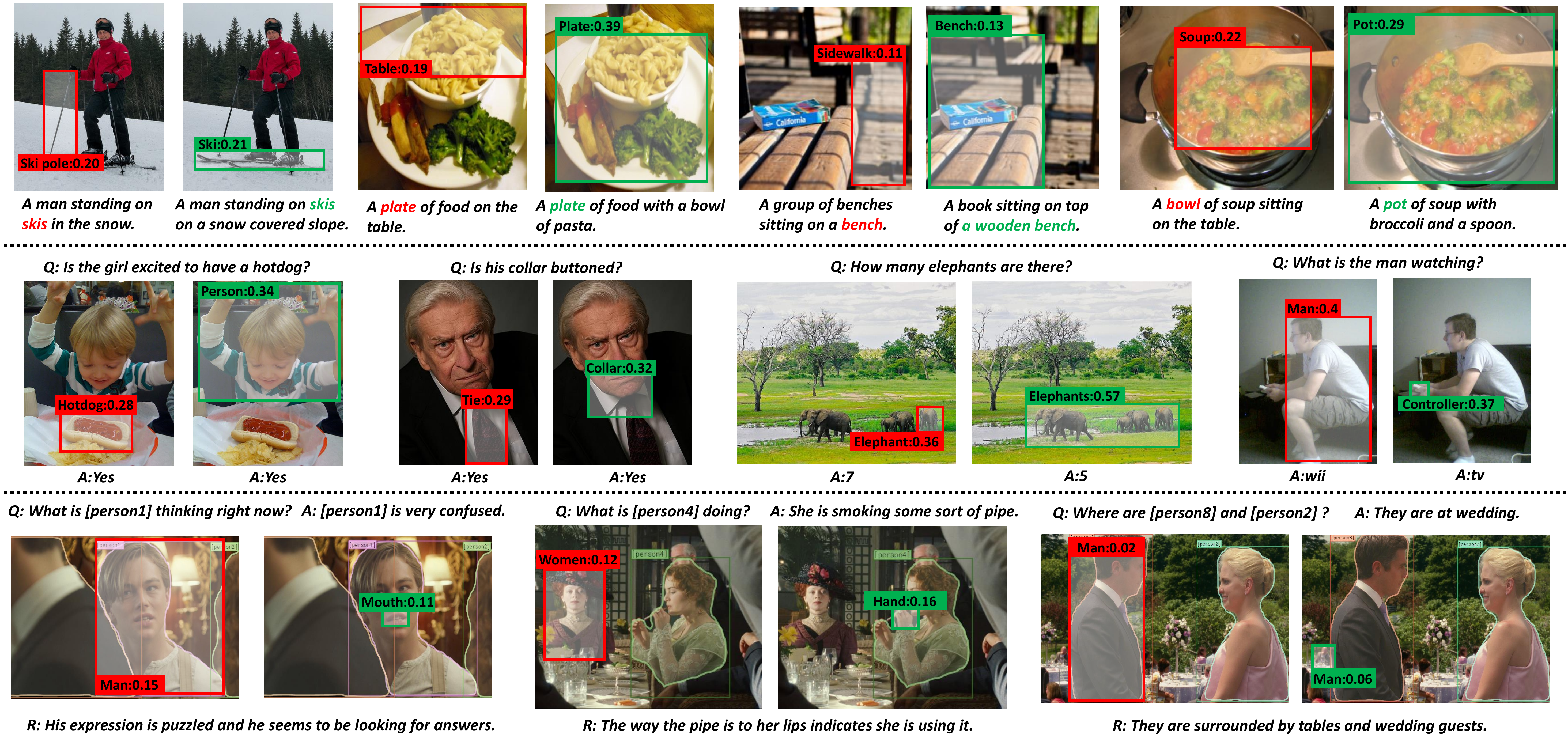}
\end{center}
\vspace{-0.3cm}
  \caption{Qualitative examples of utilizing our VC feature (right) compared with using Obj feature (left). Boxes in images denote the attention region labeled with name and attention weight. Three rows represent Image Captioning, VQA and VCR task respectively.}
\label{fig:quant}
\vspace{-0.3cm}
\end{figure*}

\subsection{Results and Analysis}
\label{sec:results}

\noindent\textbf{Results on Image Captioning.}
We compared our VC representation with ablative features on two representative approaches: Up-Down~\cite{anderson2018bottom} and AoANet~\cite{huang2019attention}.
For Up-Down model shown in Table~\ref{tab:caption}, we can observe that with our +VC trained on MS-COCO, the model can even outperform current SOTA method AoANet over most of the metrics.
However, only utilizing the pure VC feature (\ies, Only VC) would hurt the model performance. The reason can be obvious.
Even for human it is insufficient to merely know the common sense that ``apple is edible'' for specific tasks, we also need visual features containing objects and attributes (\egs, ``what color is the apple'') which are encoded by previous representations. 
When comparing +VC with the +Det and +Cor without intervention, results also show absolute gains over all metrics, which demonstrates the effectiveness of our proposed causal intervention in representation learning.
AoANet~\cite{huang2019attention} proposed an ``Attention on Attention'' module on feature encoder and caption decoder for refining with the self-attention mechanism.
In our experiment, we discarded the AoA refining encoder (\ies, AoANet$^{\dagger}$) rather than using full AoANet since the self-attentive operation on feature can be viewed as an indiscriminate correlation against our do-expression.
From Table~\ref{tab:caption} we can observe that our +VC with AoANet$^{\dagger}$ achieves a new SOTA performance.
We also evaluated our feature on the online COCO test server in Table~\ref{tab:caption_test}. We can find our model also achieves the best single-model scores across all metrics outperforming previous methods significantly.

Moreover, since the existing metrics fall short to the dataset bias, we also applied a new metric CHAIR~\cite{rohrbach2018object} to measure the object hallucination (\egs, ``hallucinate'' objects not in image). The lower is better.
As shown in Table~\ref{tab:bias}, we can see that our VC feature performs the best on both standard and CHAIR metrics, thanks to our proposed intervention that can encode the visual commonsense knowledge.

\begin{table}[]
\centering
\scalebox{0.63}{
\begin{tabular}{p{0.8cm}lcccc}
\hline\hline
\multirow{2}{*}{\large{Model}}   & \multicolumn{1}{c}{\multirow{2}{*}{\large{Feature}}} & \multicolumn{2}{c}{\textbf{MS-COCO}} & \multicolumn{2}{c}{\textbf{Open Images}} \\
\cmidrule(lr){3-4}\cmidrule(lr){5-6}
                         & \multicolumn{1}{c}{}                         & Q$\rightarrow$ \text{A}           & QA$\rightarrow$ \text{R}          & Q$\rightarrow$ \text{A}             & QA$\rightarrow$ \text{R}            \\ \hline
\multicolumn{1}{c}{\multirow{6}{*}{\rotatebox{90}{\large{R2C}}}}     & Origin~\cite{zellers2019recognition}                                     & 63.8         & 67.2         & 63.8           & 67.2           \\
                         & Obj                                 & 65.9         & 68.2         & 65.9           & 68.2           \\
                         & Only VC                                     &64.1          &66.7         & 64.3         &66.8          \\
                         & +Det                             & 66.1         & 68.5         & 66.1           & 68.3           \\
                         & +Cor                             & 66.5         & 68.9         & 66.6           & 69.1           \\
                         & \cellcolor{mygray}+VC                              & \cellcolor{mygray}67.4         & \cellcolor{mygray}69.5         & \cellcolor{mygray}67.2           & \cellcolor{mygray}69.9           \\ \hline
\multicolumn{1}{c}{\multirow{5}{*}{\rotatebox{90}{\large{ViLBERT$^{\dagger}$}}}} 
                         & Obj\textsuperscript{\ref{ftnote}}                                  & 69.1         & 69.6         & 69.1           & 69.6           \\
                         & Only VC                                     &68.8           &70.1         &68.9         &70.1          \\
                         & +Det                         & 69.2         & 69.8           & 69.1      & 69.6       \\
                         & +Cor                             & 69.3         & 69.9         & 69.2           & 70.0           \\
                         & \cellcolor{mygray}+VC                              & \cellcolor{mygray}\textbf{69.5}         & \cellcolor{mygray}70.2        & \cellcolor{mygray}\textbf{69.5}           & \cellcolor{mygray}70.3           \\ \hline
SOTA                     & ViLBERT$^{\dagger}$~\cite{lu2019vilbert}                                      & 69.3         & \textbf{71.0}         & 69.3           & \textbf{71.0}           \\ \hline\hline
\end{tabular}}
\caption{Experimental results on VCR with various visual features. ViLBERT$^{\dagger}$~\cite{lu2019vilbert} denotes ViLBERT without pretraining process.}
\label{tab:vcr}
\vspace{-0.4cm}
\end{table}

\noindent\textbf{Results on VQA.}
In Table~\ref{tab:vqa}, we applied our VC feature on classical Up-Down~\cite{anderson2018bottom} and recent state-of-the-art method MCAN~\cite{yu2019deep}.
From the results, our proposed +VC outperforms all the other ablative representations on three answer types, achieving the state-of-the-art performance.
However, compared to the image captioning, the gains on VQA with our VC feature are less significant.
The potential reason lies in the limited ability of the current question understanding, which cannot be resolved by ``visual'' common sense.
Table~\ref{tab:vqa_test} reports the single model performance of various models on both test-dev and test-standard sets.
Although our VC feature is limited by the question understanding, we still receive the absolute gains by just feature concatenation compared to previous methods with complicated module stack, which only achieves a slight improvement.

\noindent\textbf{Results on VCR.}
We present two representative methods R2C~\cite{zellers2019recognition} and ViLBERT~\cite{lu2019vilbert} in this emerging task on the validation set.
Note that as the R2C applies the ResNet backbone for residual feature extraction, here for fair comparison we 
switched it to the uniform bottom-up features. 
Moreover, for ViLBERT, since our VC features were not involved in the pretraining process on Conceptual Captions, here we utilized the ViLBERT$^{\dagger}$~\cite{lu2019vilbert} rather than the full ViLBERT model.
From the comparison with ablative visual representations in Table~\ref{tab:vcr}, our +VC feature still shows the superior performances similar to the above two tasks.

\noindent\textbf{Results on Open Images.}
To evaluate the transfer ability and flexibility of the learned visual commonsense feature, we also performed our proposed VC R-CNN on a large image detection collection.
The results can be referred to Table~\ref{tab:caption}\&\ref{tab:vqa}\&\ref{tab:vcr}.
We can see that the performances are extremely close to the VC feature trained on MS-COCO, indicating the stability of our learned semantically meaningful representation.
Moreover, while performing VCR with the dataset of movie clip, which has quite diverse distributions compared to the captioning and VQA built on MS-COCO, our VC R-CNN trained on Open Images achieves the reasonable better results.

\begin{table}[]
\centering
\scalebox{0.8}{
\begin{tabular}{lccc}
\hline\hline
Component                   & Setting            & CIDEr-D             & Accuracy             \\ \hline
Expectation                        & $\mathbb{E}_{\bm{z}}[\bm{z}]$      & 128.9               & 67.2            \\ \hline
NCC                         & w/o NCC            & 131.5               & \textbf{67.7}            \\ \hline
\multirow{3}{*}{Dictionary} & Random Dictionary  & 127.5               & 66.9              \\
                            & Context Dictionary & \multicolumn{2}{c}{\textit{Unstable Training}} \\
                            & Fixed Dictionary     & \textbf{131.6}               & \textbf{67.7}            \\ \hline\hline
\end{tabular}}
\caption{Ablation studies of our proposed intervention trained on MS-COCO and evaluated with CIDEr-D (captioning) and Accuracy (VQA) on Karpathy testset and VQA2.0 validation set.}
\label{tab:ablation}
\vspace{-0.5cm}
\end{table}

\subsection{Qualitative Analysis}
We visualize several examples with our VC feature and previous Up-Down feature~\cite{anderson2018bottom} for each task in Figure~\ref{fig:quant}. Any other settings except for feature kept the same. We can observe that with our VC, models can choose more precise, reasonable attention area and explicable better performance.

\subsection{Ablation Study}
To evaluate our proposed intervention implementation, we carry out different settings for each module in our VC R-CNN and report results on captioning and VQA in Table~\ref{tab:ablation}.
$\mathbb{E}_{\bm{z}}[\bm{z}]$ denotes utilizing statistical $P(z)$ by counting from the dataset without attention.
Random Dictionary denotes initializing the confounder dictionary by randomization rather than the average RoI feature, while the Context Dictionary encodes contexts in each image as a dynamic dictionary set.
The default setting is the fixed confounder dictionary with our attention module and NCC, which gives the best results.
We can observe that random dictionary and $\mathbb{E}_{\bm{z}}[\bm{z}]$ would hurt the performance, which demonstrates the effectiveness of our implementation.
Moreover, we can find that NCC refining just brings a little difference to the downstream task performance. The potential reason is that NCC just provides a qualitative prediction and may have deviation when applying on real-world visual feature. We will continue exploring NCC in the future work.
\vspace{-0.1cm}

\section{Conclusions}

\vspace{-0.1cm}

We presented a novel unsupervised feature representation learning method called VC R-CNN that can be based on any R-CNN framework, supporting a variety of high-level tasks by using only feature concatenation. The key novelty of VC R-CNN is that the learning objective is based on causal intervention, which is fundamentally different from the conventional likelihood. Extensive experiments on benchmarks showed impressive performance boosts on almost all the strong baselines and metrics. In future, we intend to study the potential of our VC R-CNN applied in other modalities such as video and 3D point cloud.

\noindent\textbf{Acknowledgments~}
We would like to thank all reviewers for their constructive comments. This work was partially supported by the NTU-Alibaba JRI and the Singapore Ministry of Education (MOE) Academic Research Fund (AcRF) Tier 1 grant.

{\small
\bibliographystyle{ieee_fullname}
\bibliography{egbib}

\begin{thebibliography}{10}\itemsep=-1pt

\bibitem{anderson2016spice}
Peter Anderson, Basura Fernando, Mark Johnson, and Stephen Gould.
\newblock Spice: Semantic propositional image caption evaluation.
\newblock In {\em ECCV}. Springer, 2016.

\bibitem{anderson2018bottom}
Peter Anderson, Xiaodong He, Chris Buehler, Damien Teney, Mark Johnson, Stephen
  Gould, and Lei Zhang.
\newblock Bottom-up and top-down attention for image captioning and visual
  question answering.
\newblock In {\em CVPR}, 2018.

\bibitem{antol2015vqa}
Stanislaw Antol, Aishwarya Agrawal, Jiasen Lu, Margaret Mitchell, Dhruv Batra,
  C Lawrence~Zitnick, and Devi Parikh.
\newblock Vqa: Visual question answering.
\newblock In {\em ICCV}, 2015.

\bibitem{banerjee2005meteor}
Satanjeev Banerjee and Alon Lavie.
\newblock Meteor: An automatic metric for mt evaluation with improved
  correlation with human judgments.
\newblock In {\em ACLW}, 2005.

\bibitem{bengio2019meta}
Yoshua Bengio, Tristan Deleu, Nasim Rahaman, Rosemary Ke, S{\'e}bastien
  Lachapelle, Olexa Bilaniuk, Anirudh Goyal, and Christopher Pal.
\newblock A meta-transfer objective for learning to disentangle causal
  mechanisms.
\newblock {\em arXiv preprint arXiv:1901.10912}, 2019.

\bibitem{bengio2014deep}
Yoshua Bengio, Eric Laufer, Guillaume Alain, and Jason Yosinski.
\newblock Deep generative stochastic networks trainable by backprop.
\newblock In {\em ICML}, 2014.

\bibitem{cadene2019rubi}
Remi Cadene, Corentin Dancette, Matthieu Cord, Devi Parikh, et~al.
\newblock Rubi: Reducing unimodal biases for visual question answering.
\newblock In {\em NIPS}, 2019.

\bibitem{chalupka2014visual}
Krzysztof Chalupka, Pietro Perona, and Frederick Eberhardt.
\newblock Visual causal feature learning.
\newblock {\em arXiv preprint arXiv:1412.2309}, 2014.

\bibitem{chen2019uniter}
Yen-Chun Chen, Linjie Li, Licheng Yu, Ahmed~El Kholy, Faisal Ahmed, Zhe Gan, Yu
  Cheng, and Jingjing Liu.
\newblock Uniter: Learning universal image-text representations.
\newblock {\em arXiv preprint arXiv:1909.11740}, 2019.

\bibitem{dai-etal-2019-transformer}
Zihang Dai, Zhilin Yang, Yiming Yang, Jaime Carbonell, Quoc Le, and Ruslan
  Salakhutdinov.
\newblock Transformer-{XL}: Attentive language models beyond a fixed-length
  context.
\newblock In {\em ACL}, July 2019.

\bibitem{dasgupta2019causal}
Ishita Dasgupta, Jane Wang, Silvia Chiappa, Jovana Mitrovic, Pedro Ortega,
  David Raposo, Edward Hughes, Peter Battaglia, Matthew Botvinick, and Zeb
  Kurth-Nelson.
\newblock Causal reasoning from meta-reinforcement learning.
\newblock {\em arXiv preprint arXiv:1901.08162}, 2019.

\bibitem{devlin2018bert}
Jacob Devlin, Ming-Wei Chang, Kenton Lee, and Kristina Toutanova.
\newblock {BERT}: Pre-training of deep bidirectional transformers for language
  understanding.
\newblock In {\em NAACL}, June 2019.

\bibitem{doersch2015unsupervised}
Carl Doersch, Abhinav Gupta, and Alexei~A Efros.
\newblock Unsupervised visual representation learning by context prediction.
\newblock In {\em ICCV}, 2015.

\bibitem{domke2008killed}
Justin Domke, Alap Karapurkar, and Yiannis Aloimonos.
\newblock Who killed the directed model?
\newblock In {\em CVPR}. IEEE, 2008.

\bibitem{d2019multi}
Alexander D’Amour.
\newblock On multi-cause approaches to causal inference with unobserved
  counfounding: Two cautionary failure cases and a promising alternative.
\newblock In {\em AISTATS}, 2019.

\bibitem{gao2019dynamic}
Peng Gao, Zhengkai Jiang, Haoxuan You, Pan Lu, Steven~CH Hoi, Xiaogang Wang,
  and Hongsheng Li.
\newblock Dynamic fusion with intra-and inter-modality attention flow for
  visual question answering.
\newblock In {\em CVPR}, 2019.

\bibitem{gibson1977theory}
James~J Gibson.
\newblock The theory of affordances.
\newblock {\em Hilldale, USA}, 1(2), 1977.

\bibitem{gidaris2018unsupervised}
Spyros Gidaris, Praveer Singh, and Nikos Komodakis.
\newblock Unsupervised representation learning by predicting image rotations.
\newblock In {\em ICLR}, 2018.

\bibitem{goyal2017something}
Raghav Goyal, Samira~Ebrahimi Kahou, Vincent Michalski, Joanna Materzynska,
  Susanne Westphal, Heuna Kim, Valentin Haenel, Ingo Fruend, Peter Yianilos,
  Moritz Mueller-Freitag, et~al.
\newblock The" something something" video database for learning and evaluating
  visual common sense.
\newblock In {\em ICCV}, volume~1, 2017.

\bibitem{goyal2017making}
Yash Goyal, Tejas Khot, Douglas Summers-Stay, Dhruv Batra, and Devi Parikh.
\newblock Making the v in vqa matter: Elevating the role of image understanding
  in visual question answering.
\newblock In {\em CVPR}, 2017.

\bibitem{halloun1985common}
Ibrahim~Abou Halloun and David Hestenes.
\newblock Common sense concepts about motion.
\newblock {\em American journal of physics}, 53(11), 1985.

\bibitem{he2017mask}
Kaiming He, Georgia Gkioxari, Piotr Doll{\'a}r, and Ross Girshick.
\newblock Mask r-cnn.
\newblock In {\em ICCV}, 2017.

\bibitem{he2016deep}
Kaiming He, Xiangyu Zhang, Shaoqing Ren, and Jian Sun.
\newblock Deep residual learning for image recognition.
\newblock In {\em CVPR}, 2016.

\bibitem{hendricks2018women}
Lisa~Anne Hendricks, Kaylee Burns, Kate Saenko, Trevor Darrell, and Anna
  Rohrbach.
\newblock Women also snowboard: Overcoming bias in captioning models.
\newblock In {\em ECCV}. Springer, 2018.

\bibitem{huang2019attention}
Lun Huang, Wenmin Wang, Jie Chen, and Xiao-Yong Wei.
\newblock Attention on attention for image captioning.
\newblock In {\em ICCV}, 2019.

\bibitem{kalainathan2018sam}
Diviyan Kalainathan, Olivier Goudet, Isabelle Guyon, David Lopez-Paz, and
  Mich{\`e}le Sebag.
\newblock Sam: Structural agnostic model, causal discovery and penalized
  adversarial learning.
\newblock {\em arXiv preprint arXiv:1803.04929}, 2018.

\bibitem{Kim2018}
Jin-Hwa Kim, Jaehyun Jun, and Byoung-Tak Zhang.
\newblock {Bilinear Attention Networks}.
\newblock In {\em NIPS}, 2018.

\bibitem{kocaoglu2017causalgan}
Murat Kocaoglu, Christopher Snyder, Alexandros~G. Dimakis, and Sriram
  Vishwanath.
\newblock Causalgan: Learning causal implicit generative models with
  adversarial training.
\newblock In {\em ICLR}, 2018.

\bibitem{kolesnikov2019revisiting}
Alexander Kolesnikov, Xiaohua Zhai, and Lucas Beyer.
\newblock Revisiting self-supervised visual representation learning.
\newblock In {\em CVPR}, 2019.

\bibitem{kristan2015visual}
Matej Kristan, Jiri Matas, Ales Leonardis, Michael Felsberg, Luka Cehovin,
  Gustavo Fernandez, Tomas Vojir, Gustav Hager, Georg Nebehay, and Roman
  Pflugfelder.
\newblock The visual object tracking vot2015 challenge results.
\newblock In {\em ICCVW}, 2015.

\bibitem{krizhevsky2012imagenet}
Alex Krizhevsky, Ilya Sutskever, and Geoffrey~E Hinton.
\newblock Imagenet classification with deep convolutional neural networks.
\newblock In {\em NIPS}, 2012.

\bibitem{kuznetsova2018open}
Alina Kuznetsova, Hassan Rom, Neil Alldrin, Jasper Uijlings, Ivan Krasin, Jordi
  Pont-Tuset, Shahab Kamali, Stefan Popov, Matteo Malloci, Alexander
  Kolesnikov, et~al.
\newblock The open images dataset v4.
\newblock {\em IJCV}, 2020.

\bibitem{lee2019rethinking}
Hankook Lee, Sung~Ju Hwang, and Jinwoo Shin.
\newblock Rethinking data augmentation: Self-supervision and self-distillation.
\newblock {\em arXiv preprint arXiv:1910.05872}, 2019.

\bibitem{li2019siamrpn++}
Bo Li, Wei Wu, Qiang Wang, Fangyi Zhang, Junliang Xing, and Junjie Yan.
\newblock Siamrpn++: Evolution of siamese visual tracking with very deep
  networks.
\newblock In {\em CVPR}, 2019.

\bibitem{lin2004rouge}
Chin-Yew Lin.
\newblock Rouge: A package for automatic evaluation of summaries.
\newblock In {\em Text summarization branches out}, 2004.

\bibitem{lin2014microsoft}
Tsung-Yi Lin, Michael Maire, Serge Belongie, James Hays, Pietro Perona, Deva
  Ramanan, Piotr Doll{\'a}r, and C~Lawrence Zitnick.
\newblock Microsoft coco: Common objects in context.
\newblock In {\em ECCV}. Springer, 2014.

\bibitem{lin2015don}
Xiao Lin and Devi Parikh.
\newblock Don't just listen, use your imagination: Leveraging visual common
  sense for non-visual tasks.
\newblock In {\em CVPR}, 2015.

\bibitem{liu2016ssd}
Wei Liu, Dragomir Anguelov, Dumitru Erhan, Christian Szegedy, Scott Reed,
  Cheng-Yang Fu, and Alexander~C Berg.
\newblock Ssd: Single shot multibox detector.
\newblock In {\em ECCV}, 2016.

\bibitem{long2015fully}
Jonathan Long, Evan Shelhamer, and Trevor Darrell.
\newblock Fully convolutional networks for semantic segmentation.
\newblock In {\em CVPR}, 2015.

\bibitem{lopez2017discovering}
David Lopez-Paz, Robert Nishihara, Soumith Chintala, Bernhard Scholkopf, and
  L{\'e}on Bottou.
\newblock Discovering causal signals in images.
\newblock In {\em CVPR}, 2017.

\bibitem{lu2019vilbert}
Jiasen Lu, Dhruv Batra, Devi Parikh, and Stefan Lee.
\newblock Vilbert: Pretraining task-agnostic visiolinguistic representations
  for vision-and-language tasks.
\newblock In {\em NIPS}, 2019.

\bibitem{maaten2008visualizing}
Laurens van~der Maaten and Geoffrey Hinton.
\newblock Visualizing data using t-sne.
\newblock {\em JMLR}, 9(Nov), 2008.

\bibitem{malisiewicz2009beyond}
Tomasz Malisiewicz and Alyosha Efros.
\newblock Beyond categories: The visual memex model for reasoning about object
  relationships.
\newblock In {\em NIPS}, 2009.

\bibitem{manjunatha2019explicit}
Varun Manjunatha, Nirat Saini, and Larry~S Davis.
\newblock Explicit bias discovery in visual question answering models.
\newblock In {\em CVPR}, 2019.

\bibitem{mikolov2013distributed}
Tomas Mikolov, Ilya Sutskever, Kai Chen, Greg~S Corrado, and Jeff Dean.
\newblock Distributed representations of words and phrases and their
  compositionality.
\newblock In {\em NIPS}, 2013.

\bibitem{nair2019causal}
Suraj Nair, Yuke Zhu, Silvio Savarese, and Li Fei-Fei.
\newblock Causal induction from visual observations for goal directed tasks.
\newblock {\em arXiv preprint arXiv:1910.01751}, 2019.

\bibitem{papineni2002bleu}
Kishore Papineni, Salim Roukos, Todd Ward, and Wei-Jing Zhu.
\newblock Bleu: a method for automatic evaluation of machine translation.
\newblock In {\em ACL}. Association for Computational Linguistics, 2002.

\bibitem{pearl2014interpretation}
Judea Pearl.
\newblock Interpretation and identification of causal mediation.
\newblock {\em Psychological methods}, 19(4), 2014.

\bibitem{pearl2016causal}
Judea Pearl, Madelyn Glymour, and Nicholas~P Jewell.
\newblock {\em Causal inference in statistics: A primer}.
\newblock John Wiley \& Sons, 2016.

\bibitem{pearl2018book}
Judea Pearl and Dana Mackenzie.
\newblock {\em The book of why: the new science of cause and effect}.
\newblock Basic Books, 2018.

\bibitem{peters2018deep}
Matthew Peters, Mark Neumann, Mohit Iyyer, Matt Gardner, Christopher Clark,
  Kenton Lee, and Luke Zettlemoyer.
\newblock Deep contextualized word representations.
\newblock In {\em NAACL}, 2018.

\bibitem{qi2019two}
Jiaxin Qi, Yulei Niu, Jianqiang Huang, and Hanwang Zhang.
\newblock Two causal principles for improving visual dialog.
\newblock In {\em CVPR}, 2020.

\bibitem{ramakrishnan2018overcoming}
Sainandan Ramakrishnan, Aishwarya Agrawal, and Stefan Lee.
\newblock Overcoming language priors in visual question answering with
  adversarial regularization.
\newblock In {\em NIPS}, 2018.

\bibitem{ren2015faster}
Shaoqing Ren, Kaiming He, Ross Girshick, and Jian Sun.
\newblock Faster r-cnn: Towards real-time object detection with region proposal
  networks.
\newblock In {\em NIPS}, 2015.

\bibitem{rohrbach2018object}
Anna Rohrbach, Lisa~Anne Hendricks, Kaylee Burns, Trevor Darrell, and Kate
  Saenko.
\newblock Object hallucination in image captioning.
\newblock In {\em EMNLP}, 2018.

\bibitem{rosenfeld2011common}
Sophia~A Rosenfeld.
\newblock {\em Common sense}.
\newblock Harvard University Press, 2011.

\bibitem{sadeghi2015viske}
Fereshteh Sadeghi, Santosh~K Kumar~Divvala, and Ali Farhadi.
\newblock Viske: Visual knowledge extraction and question answering by visual
  verification of relation phrases.
\newblock In {\em CVPR}, 2015.

\bibitem{smith1995structures}
Barry Smith.
\newblock The structures of the common-sense world.
\newblock 1995.

\bibitem{su2018learning}
Zhou Su, Chen Zhu, Yinpeng Dong, Dongqi Cai, Yurong Chen, and Jianguo Li.
\newblock Learning visual knowledge memory networks for visual question
  answering.
\newblock In {\em CVPR}, 2018.

\bibitem{sun2019videobert}
Chen Sun, Austin Myers, Carl Vondrick, Kevin Murphy, and Cordelia Schmid.
\newblock Videobert: A joint model for video and language representation
  learning.
\newblock In {\em ICCV}, 2019.

\bibitem{tan2019lxmert}
Hao Tan and Mohit Bansal.
\newblock {LXMERT}: Learning cross-modality encoder representations from
  transformers.
\newblock In {\em EMNLP-IJCNLP}, Nov. 2019.

\bibitem{tang2020unbiased}
Kaihua Tang, Yulei Niu, Jianqiang Huang, Jiaxin Shi, and Hanwang Zhang.
\newblock Unbiased scene graph generation from biased training.
\newblock In {\em CVPR}, 2020.

\bibitem{theis2015generative}
Lucas Theis and Matthias Bethge.
\newblock Generative image modeling using spatial lstms.
\newblock In {\em NIPS}, 2015.

\bibitem{vedantam2015cider}
Ramakrishna Vedantam, C Lawrence~Zitnick, and Devi Parikh.
\newblock Cider: Consensus-based image description evaluation.
\newblock In {\em CVPR}, 2015.

\bibitem{vedantam2015learning}
Ramakrishna Vedantam, Xiao Lin, Tanmay Batra, C Lawrence~Zitnick, and Devi
  Parikh.
\newblock Learning common sense through visual abstraction.
\newblock In {\em ICCV}, 2015.

\bibitem{vincent2008extracting}
Pascal Vincent, Hugo Larochelle, Yoshua Bengio, and Pierre-Antoine Manzagol.
\newblock Extracting and composing robust features with denoising autoencoders.
\newblock In {\em ICML}. ACM, 2008.

\bibitem{vinyals2015show}
Oriol Vinyals, Alexander Toshev, Samy Bengio, and Dumitru Erhan.
\newblock Show and tell: A neural image caption generator.
\newblock In {\em CVPR}, 2015.

\bibitem{wu2016ask}
Qi Wu, Peng Wang, Chunhua Shen, Anthony Dick, and Anton van~den Hengel.
\newblock Ask me anything: Free-form visual question answering based on
  knowledge from external sources.
\newblock In {\em CVPR}, 2016.

\bibitem{xu2015show}
Kelvin Xu, Jimmy Ba, Ryan Kiros, Kyunghyun Cho, Aaron Courville, Ruslan
  Salakhudinov, Rich Zemel, and Yoshua Bengio.
\newblock Show, attend and tell: Neural image caption generation with visual
  attention.
\newblock In {\em ICML}, 2015.

\bibitem{yang2019auto}
Xu Yang, Kaihua Tang, Hanwang Zhang, and Jianfei Cai.
\newblock Auto-encoding scene graphs for image captioning.
\newblock In {\em CVPR}, 2019.

\bibitem{yang2019learning}
Xu Yang, Hanwang Zhang, and Jianfei Cai.
\newblock Learning to collocate neural modules for image captioning.
\newblock In {\em ICCV}, 2019.

\bibitem{yang2020deconfounded}
Xu Yang, Hanwang Zhang, and Jianfei Cai.
\newblock Deconfounded image captioning: A causal retrospect.
\newblock {\em arXiv preprint arXiv:2003.03923}, 2020.

\bibitem{yatskar2016stating}
Mark Yatskar, Vicente Ordonez, and Ali Farhadi.
\newblock Stating the obvious: Extracting visual common sense knowledge.
\newblock In {\em NAACL}, 2016.

\bibitem{yu2019deep}
Zhou Yu, Jun Yu, Yuhao Cui, Dacheng Tao, and Qi Tian.
\newblock Deep modular co-attention networks for visual question answering.
\newblock In {\em CVPR}, 2019.

\bibitem{zellers2019recognition}
Rowan Zellers, Yonatan Bisk, Ali Farhadi, and Yejin Choi.
\newblock From recognition to cognition: Visual commonsense reasoning.
\newblock In {\em CVPR}, 2019.

\bibitem{zhai2019s4l}
Xiaohua Zhai, Avital Oliver, Alexander Kolesnikov, and Lucas Beyer.
\newblock S4l: Self-supervised semi-supervised learning.
\newblock In {\em ICCV}, 2019.

\bibitem{zhu2014reasoning}
Yuke Zhu, Alireza Fathi, and Li Fei-Fei.
\newblock Reasoning about object affordances in a knowledge base
  representation.
\newblock In {\em ECCV}. Springer, 2014.

\end{thebibliography}
}

\clearpage



\section*{Supplementary material}

# Supplementary Material

In this Supplementary Material, we will further detail the following aspects omitted in the main paper. The detailed code guide and VC features can be referred to https://github.com/Wangt-CN/VC-R-CNN.

- Section A: the detailed derivation of the intervention in Section 3.1 Causal Intervention of the main paper.
- Section B: The details of our proposed implementation in Section 3.2 of the main paper.
- Section C: The details of the network architecture of our VC R-CNN in Section 4 in the main paper.
- Section D: more quantitative results of VC features concatenated on on different Faster R-CNN based representations.
- Section E: more qualitative visualizations compared our VC features with previous bottom-up representations [1].

## A. The Do-Expression

In our main paper, we give the do-expression Eq. (2) comparing with the Bayes rule in an intuitive way for easier understanding. In this section, we further formally explain and prove the intervention (do calculus) in causal theory which is applied in our VC R-CNN.

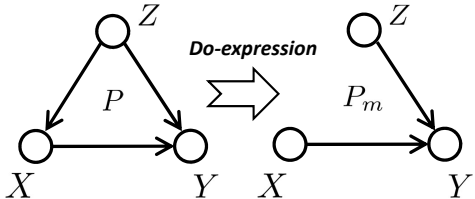

Figure 1. The do expression $P(Y|do(X))$ with a graph surgery. Nodes denote variables and arrows mean the direct causal effects.

As written in our main paper, in our visual world there may exists many "background factors" $z \in Z$, no matter known or unknown, that affect (or cause) either $X$ or/and $Y$, leading to spurious correlations by only learning from the likelihood $P(Y|X)$. To avoid the confounder as shown in Figure 1, the causal intervention (do calculus) is achieved by cutting off the effect from $Z$ to $X$ in the form of a graph surgery. Here for clear clarification, we use $P$ and $P_m$ to distinguish the probabilities in the causal graph before and after surgery, respectively. Therefore, due to the definition of the Do-expression we can have:

$$P(Y|do(X)) = P_m(Y|X). \quad \textit{(Definition)} \quad (1)$$

Then the key to compute the causal effect lies in the observation $P_m$, the manipulated probability, shares two essential properties with $P$ (*i.e.*, the original probability function that prevails in the preintervention model). First, the marginal probability $P(Z = z)$ is invariant under the intervention, because the process determining $Z$ is not affected by removing the arrow from $Z$ to $X$, *i.e.*, $P(z) = P_m(z)$. Second, the conditional probability $P(Y|X, z)$ is invariant, because the process by which $Y$ responds to $X$ and $Z$ remains the same, regardless of whether $X$ changes spontaneously or by deliberate manipulation:

$$P_m(Y|X, z) = P(Y|X, z). \quad \textit{(Invariance)} \quad (2)$$

Moreover, we can also use the fact that $Z$ and $X$ are independent under the intervention distribution. This tell us that $P_m(z|X) = P_m(z)$. Putting these considerations together, we have:

$$\begin{aligned} P(Y|do(X)) &= P_m(Y|X) \\ &= \sum_z P_m(Y|X, z) P_m(z|X) \quad \textit{(Bayes Rule)} \\ &= \sum_z P_m(Y|X, z) P_m(z) \quad \textit{(Independency)} \\ &= \sum_z P(Y|X, z) P(z), \end{aligned} \quad (3)$$

where $P(Y|X, z)$ denotes the conditional probability given $X$ and confounder $z$ and $P(z)$ is a prior probability of each object class.

The Eq. (3) is called the adjustment formula, which computed the association between $X$ and $Y$ for each value $z$ of $Z$, then averages over all values. This procedure is referred to as "adjusting for $Z$" or "controlling for $Z$". Then with this final expression, we can measure the casual effects between $X$ to $Y$ directly from the data, since it consists only of conditional probabilities.

Moreover, in the main paper to show the difference between Bayes Rule and Intervention clearly, we propose an example about `person` and `toilet` by comparing $P(Z)$ and $P(Z|\text{toilet})$ on partial labels. Here in the Supplementary Material we present the integrated figure for whole 80 MS-COCO labels on both $P(Z)$, $P(Z|X)$ and $P(Y|X, Z)P(Z), P(Y|X, Z)P(Z|X)$ in Figure 2 & 3. From Figure 2 we can see that the do intervention achieves "borrow" and "put" by applying $P(Z)$ to replace $P(Z|X)$, which can be also regarded as a kind of method to alleviate the previous long tail distribution (blue line).

## B. Our Proposed Implementation

### B.1. Normalized Weighted Geometric Mean.

In our main paper we just give the application of Normalized Weighted Geometric Mean (NWGM) due to the limited space, here we present the detailed derivation and reader can

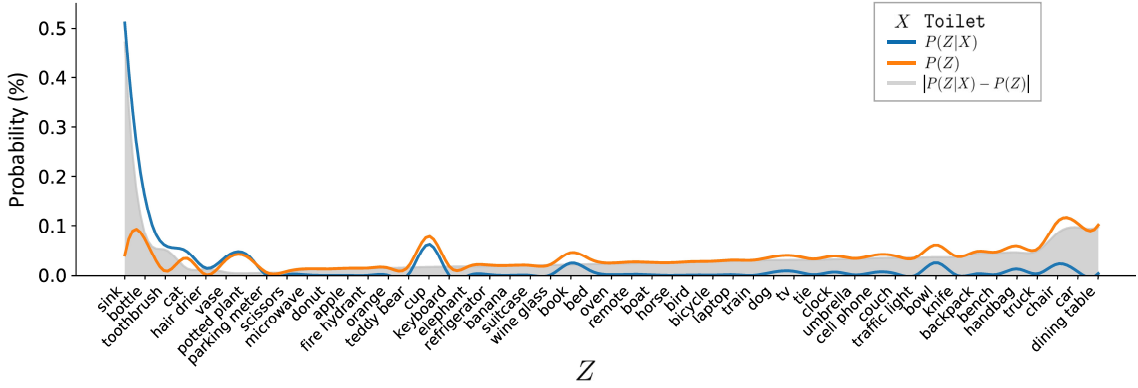

Figure 2. The case study of the differences between $P(z|\texttt{Toilet})$ and $P(z)$ from whole MS-COCO ground-truth object labels. Note that confounders that never appeared with $X$ (*i.e.*, `Toilet`) is not contained.

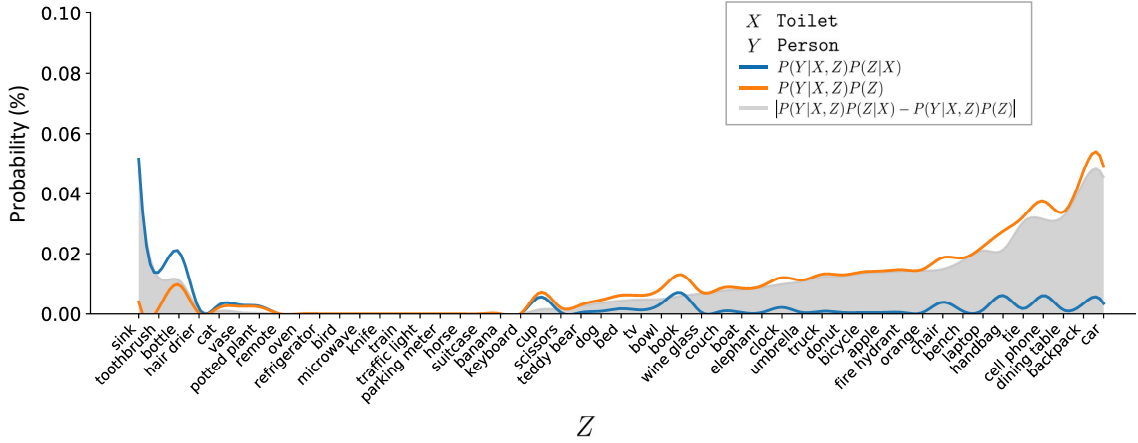

Figure 3. The case study of the differences between $P(\texttt{Person}|\texttt{Toilet},z)P(z|\texttt{Toilet})$ and $P(\texttt{Person}|\texttt{Toilet},z)P(z)$ from whole MS-COCO ground-truth object labels. Note that confounders that never appeared with $X$ (*i.e.*, `Toilet`) is not contained.

also refer to the [8]. Recall that in the main paper we have defined the RoI feature $x$ as the $X$, one of its context class label $y^c$ as $Y$. For the confounder set $Z$, we denote it as a global confounder dictionary $\boldsymbol{Z} = [\boldsymbol{z_1}, ..., \boldsymbol{z_N}]$ in the shape of $N \times d$ matrix for practical use, where $N$ is the category size in dataset (*e.g.*, 80 in MS-COCO) and $d$ is the feature dimension of $x$.

Here we first introduce the normalized weighted geometric mean in our softmax class label prediction:

$$\begin{aligned} \text{NWGM}[f_y(\boldsymbol{x}, \boldsymbol{z})] &= \frac{\prod_{\boldsymbol{z}} \exp(f_y(\boldsymbol{x},\boldsymbol{z}))^{p(\boldsymbol{z})}}{\sum_j \prod_{\boldsymbol{z}} \exp(f_y(\boldsymbol{x},\boldsymbol{z}))^{p(\boldsymbol{z})}} \\ &= \frac{\exp(\mathbb{E}_{\boldsymbol{z}}[f_y(\boldsymbol{x},\boldsymbol{z})])}{\sum_j \exp(\mathbb{E}_{\boldsymbol{z}}[f_y(\boldsymbol{x},\boldsymbol{z})])} \\ &= \textit{Softmax}(\mathbb{E}_{\boldsymbol{z}}[f_y(\boldsymbol{x},\boldsymbol{z})]), \end{aligned} \quad (4)$$

where $f_y(\cdot)$ calculates the logits for $N$ categories. Note that the subscript $y$ denotes that $f(\cdot)$ is parameterized by feature $\boldsymbol{y}$, motivated by the heuristics that the context prediction task for RoI $Y$ is characterized by its visual feature. We can see that the most ingenious operation in Eq. (4) is to change the production $\prod$ to the sum $\sum$ by putting it into the exp. Moreover, from the results in [8, 2, 7], we know $\text{NWGM}[f_y(\boldsymbol{x}, \boldsymbol{z})] \approx \mathbb{E}_{\boldsymbol{z}}[\textit{Softmax}(f_y(\boldsymbol{x},\boldsymbol{z}))]$ under the softmax activation. Therefore Eq. (3) in the main paper can be further derived as:

$$P(Y|do(X)) \approx \textit{Softmax}(\mathbb{E}_{\boldsymbol{z}}[f_y(\boldsymbol{x},\boldsymbol{z})]). \quad (5)$$

Furthermore, we use the linear model $f_y(\boldsymbol{x}, \boldsymbol{z}) = \boldsymbol{W}_1 \boldsymbol{x} + \boldsymbol{W}_2 \cdot g_y(\boldsymbol{z})$, where $\boldsymbol{W}_1, \boldsymbol{W}_2 \in \mathbb{R}^{N \times d}$ denote the fully connected layer. Then the linear projection of the expectation of one variable equals to the linear projection of that and we can put $\mathbb{E}$ into the linear projection as $\textit{Softmax}(\boldsymbol{W}_1 \mathbb{E}_{\boldsymbol{z}}[\boldsymbol{x}] + \boldsymbol{W}_2 \cdot \mathbb{E}_{\boldsymbol{z}}[g_y(\boldsymbol{z})])$. Since the RoI representation $\boldsymbol{x}$ remains the same, we can discard the $\mathbb{E}$ over $\boldsymbol{x}$, *i.e.*, $\textit{Softmax}(\boldsymbol{W}_1 \boldsymbol{x} + \boldsymbol{W}_2 \cdot \mathbb{E}_{\boldsymbol{z}}[g_y(\boldsymbol{z})])$. That means the expectation of the outputs over all possible confounder $\boldsymbol{z}$ can be simply computed by feedforward propagation with

| Index | Input | Operation | Output | Trainable Parameters |
|---|---|---|---|---|
| (1) | - | RoI feature | $\boldsymbol{x}$ ($1024 \times 1$) | - |
| (2) | - | RoI feature | $\boldsymbol{y}$ ($1024 \times 1$) | - |
| (3) | (2), $\boldsymbol{Z}$ | Scale Dot-Product Attention | $\mathbb{E}_{\boldsymbol{z}}[g_y(\boldsymbol{z})]$ ($1024 \times 1$) | $\boldsymbol{W}_3$ ($512 \times 1024$) $\boldsymbol{W}_4$ ($512 \times 1024$) |
| (4) | (1),(3) | Linear Addition Model | $\mathbb{E}_{\boldsymbol{z}}[f_y(\boldsymbol{x}, \boldsymbol{z})]$ ($80 \times 1$) | $\boldsymbol{W}_1$ ($80 \times 1024$) $\boldsymbol{W}_2$ ($80 \times 1024$) |
| (5) | (1) | Feature Embedding | $\boldsymbol{W}\boldsymbol{x}$ ($80 \times 1$) | $\boldsymbol{W}$ ($80 \times 1024$) |
| (6) | (5) | Self Predictor | *Softmax* | - |
| (7) | (4) | Context Predictor | *Softmax* | - |

Table 1. The detailed network architecture of our VC R-CNN.

the expectation vector $\mathbb{E}_{\boldsymbol{z}}[g_y(\boldsymbol{z})]$ as the input.

## B.2. Neural Causation Coefficient (NCC)

Here we give a more detailed information about the usage of NCC and collider of our proposed implementations in the main paper. In our visual world sometimes there are no confounders in the structure like $X \rightarrow Z \leftarrow Y$ what we call "collider", as shown in Figure 4. Felix Elwert and Chris Winship [4] have illustrated this junction using three features of Hollywood actors: Talent ($X$), Celebrity ($Z$), and Beauty ($Y$). Here we are asserting that both talent and beauty contribute to an actor's success, but beauty and talent are completely unrelated to one another in the general population.

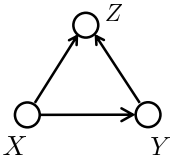

Figure 4. The causal graph structure of the "collider". Nodes denote variables, arrows denote the direct causal effects.

In this structure making the intervention on variable $Z$ (*i.e.*, condition on $Z$) would create a spurious dependence between $X$ and $Y$. The reason is that if $X$ and $Y$ are independent to begin with, conditioning on $Z$ will make them dependent. For example, if we look only at famous actors (in other words, we observe the variable Celebrity = 1), we will see a negative correlation between talent and beauty: finding out that a celebrity is unattractive increases our belief that he or she is talented. This negative correlation is sometimes called collider bias or the "explain-away" effect. Therefore we cannot make the intervention as what we do before in the collider structure. For simplicity we would make a preliminary examination before training to eliminate the effect of collider in the whole dataset. We apply the neural causation inference model (NCC) [6] to detect the strong causal effect from $X \rightarrow Z$ and $Y \rightarrow Z$ with the RoI feature directly.

NCC has partly proven [6] to be efficient for transferring to real-world, visual cause-effect observational samples with just training on artificially constructed synthetic observational samples. Specifically, the $n$ synthetic observational samples $S_i = \{(x_{ij}, y_{ij})\}_{j=1}^{m_i}$ are drawn from an heteroscedastic additive noise model $y_{ij} = f_i(x_{ij}) + v_{ij} e_{ij}$ for all $j = 1, ..., m_i$. The cause terms $x_{ij}$ are drawn from a mixture of $k_i$ Gaussians distributions. We construct each Gaussian by sampling its mean from Gaussian(0, $r_i$), its standard deviation from Gaussian(0, $s_i$) followed by an absolute value, and its unnormalized mixture weight from Gaussian (0, 1) followed by an absolute value. NCC samples $k_i$ from RandomInteger[1,5] and $r_i$, $s_i$ from Uniform[0,5]. NCC normalizes the mixture weights to sum to one and $x_{ij}{}_{j=1}^{m_i}$ to zero mean and unit variance. The noise term $v_{ij}$ and $e_{ij}$ are also sampled from Gaussian distribution and mechanism $f_i$ is a cubic hermite spline which can be referred to [6]. Finally NCC is trained with two embedding layers and two classification layers followed by the softmax in a ternary classification task (causal, anticausal and no causation). Then while testing the model can be used to evaluate on the RoI feature vectors directly. The output $NCC(\boldsymbol{x} \rightarrow \boldsymbol{y})$ ranges from $(0, 1)$ denotes the relative causality intensity from $\boldsymbol{x}$ inferring $\boldsymbol{y}$.

However since the NCC model just can provide a qualitative prediction and may have huge deviation when applying on real-world feature which may affects the training procedure of our VC R-CNN, in our experiment we just discard few training samples with very strong collider causal structure (*i.e.*, $X \rightarrow Z \leftarrow Y$) by setting a threshold (we set 0.001 in our experiment). Moreover, we use the object-level RoI features extracted by the pretrained Faster R-CNN to pre-calculate the NCC score, which may also lead to a deviation since the pretrained RoI representations may not fully present the objects. From the Table 7 in the main paper we can also observe that NCC refining just brings a little difference to the downstream task performance. The potential reason is that our VC R-CNN can automatically learn the reasonable confounder attention during the large dataset training. We will continue exploring the usage of NCC and other causal discovery method in our future work.

| Model | Feature | Cross-Entropy Loss | | | | | | CIDEr Optimization | | | | | |
|---|---|---|---|---|---|---|---|---|---|---|---|---|---|
| | | B@1 | B@4 | M | R | S | C | B@1 | B@4 | M | R | S | C |
| Up-Down | Obj | 74.5 | 33.2 | 25.9 | 54.7 | 18.9 | 104.7 | 77.1 | 32.6 | 25.2 | 55.2 | 18.3 | 110.6 |
| | Obj+Det | 75.4 | 34.4 | 26 | 55.8 | 19.9 | 108.9 | 77.9 | 33.9 | 25.4 | 56.1 | 19.8 | 114.7 |
| | Obj+Cor | 75.6 | 34.5 | 26.1 | 55.2 | 19.6 | 108.7 | 78.0 | 34.1 | 25.6 | 56.0 | 19.9 | 115.2 |
| | Obj+VC | 76.3 | 35.3 | 26.3 | 56.3 | 20.2 | 111.6 | 79.1 | 35.7 | 25.9 | 57.0 | 20.5 | 119.7 |
| AoANet | Obj | 74.6 | 34.1 | 25.9 | 55.4 | 19.7 | 108.1 | 78.1 | 35.4 | 25.6 | 56.7 | 20.7 | 118.4 |
| | Obj+Det | 75.1 | 33.9 | 26.1 | 55.7 | 19.8 | 109.7 | 78.3 | 36.2 | 27.1 | 56.9 | 20.9 | 120.2 |
| | Obj+Cor | 75.5 | 34.3 | 26.2 | 55.9 | 20.1 | 110.8 | 78.7 | 36.8 | 27.5 | 57.2 | 21.1 | 121.1 |
| | Obj+VC | 76.0 | 35.0 | 26.4 | 56.1 | 20.5 | 112.2 | 79.1 | 37.2 | 29.0 | 57.6 | 21.5 | 123.5 |

Table 2. The image captioning performances of two models with ablative features (based on vanilla Faster R-CNN feature) on Karpathy split.

## C. Network Architecture

Here we introduce the detailed network architectures of all the components of our VC R-CNN in Table 1. Given an image and the feature extraction backbone, any two RoI feature vectors $x$ and $y$ were extracted as in Table 1 (1)(2). Then as the Section 3.2 The Proposed Implementation, we adopted the Scale Dot-Product Attention to refine confounders from the confounder dictionary $Z$ as in Table 1 (3). A linear addition model $f_y(x, z)$ was proposed to combine the effect on $Y$ from both $X$ and confounder $Z$. Finally we made the do calculus by Self Predictor and Context Predictor in Table 1 (6)(7).

## D. More Quantitative Results

In the experiment of our main paper, we adopted the bottom-up feature [1] as our base feature. The bottom-up feature pretrained Faster R-CNN on ImageNet [3] and Visual Genome [5] to propose salient object level features with attribute rather than the uniform gird of equally-sized image regions, enable attention to be calculated at the level of semantically meaningful regions and bring a huge improvement in image-and-language tasks.

Here we also concatenated our VC feature onto the vanilla image region representations based on pretrained Faster R-CNN model with ResNet-101 on MS-COCO dataset. Note that for better comparison we utilized the bounding box coordinates of the bottom-up feature to control the number and location of the boxes and then applied new feature in the Image Captioning task. Results are shown in Table 2. We can also observe that concatenating with our VC feature can lead to a huge performance improvement, which demonstrates the stability of our VC feature and effectiveness of the proposed intervention.

## E. More Qualitative Results

### E.1. Failure Case

**Failure in VC R-CNN.** As shown in Figure 5, we can see that sometimes our VC R-CNN cannot make quite reasonable refinement for confounder dictionary via the Scaled

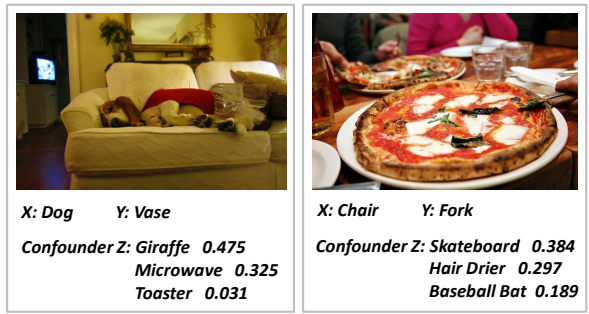

Figure 5. The examples of the failure case about confounder finding in VC R-CNN.

Dot-Product Attention while predicting $Y$ given $X$ and $Z$, especially when there is no obvious relation between $X$ and $Y$. For example while making the intervention between `dog` and `vase`, `chair` and `fork`, the model attends to the `giraffe` and `skateboard` respectively. To tackle this limitation, the better schedule of confounder exploring, for example choosing approapiate context objects as the confounder dictionary, will be tried in our future work.

**Failure in Downstream Tasks.** Though we designed the intervention (do-expression) in unsupervised representation learning to prevent the cognition error and help machine learn the common sense, some attention errors still exist in downstream tasks. Here we present two examples in Figure 6. We can observe that in the VQA example (left), the model provides a reasonable but incorrect answer, while in image captioning the generated description does not cover every instance. The possible reason lies in two folds. First, the current detection technique is still limited, for example the Faster R-CNN cannot recognize the `kangaroo` on the stop sign. Second, we know that our VC R-CNN can find the probable and reasonable confounders from the confounder dictionary according to the given image. However, it may still fail to exploit the exact confounder (*e.g.*, `motorcycle` in VQA and `lamp`, `chair` in Image Captioning) to fully eliminate the correlation bias.

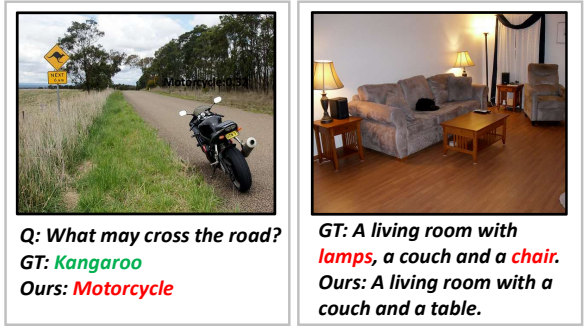

Figure 6. The examples of the failure case in downstream tasks.

### E.2. Image Captioning

Figure 7 & 8 exhibit visualizations of utilizing our VC feature (right) compared with using Faster R-CNN feature (*i.e.*, bottom-up feature, left) with the classical Up-Down model in image captioning task. The boxes represent the attended regions when generating words with the same color. From the illustration we can observe that with our VC feature, model can generate more fruitful descriptions with more accurate attention. For example, in Figure 8 bottom with our VC feature, model focuses on birds and gives the accurate and fruitful descriptions: "two birds perched" rather than "a bird sitting" generated by the baseline model. Furthermore, we can also see that our VC feature can help to overcome the language bias efficiently. Other than giving the common collections, the model can generate reasonable captions according to the image content. For example in the middle of Figure 8, "cat" appearing with "bed" ("cat+bed"/"cat"=6.7%) is quite more often than "cat" with "blanket" ("cat+blanket"/"cat"=1.4%) in the training text, leading to a "hallucination" to generate "bed" without seeing the bed.

### E.3. VQA

We presented the comparison of Faster R-CNN feature (left) and our VC feature (right) in VQA in Figure 9 & 10 based on the Up-Down model. We can see that in VQA task the most serious problem is the incorrect attention even with the correct answer, which means the model actually NOT understand the question and make inference combining the vision and language. As we described in Introduction of the main paper, the dataset co-occurring bias may lead to the incorrect attention. For example in the middle of Figure 9 the model attend to the horse rather than human since horse and person co-occur too many times. Thanks to our VC feature, the attention becomes better and more accurate with alleviating the correlation bias by our proposed intervention.

## References

[1] Peter Anderson, Xiaodong He, Chris Buehler, Damien Teney, Mark Johnson, Stephen Gould, and Lei Zhang. Bottom-up and top-down attention for image captioning and visual question answering. In *CVPR*, pages 6077–6086, 2018. 1, 4

[2] Pierre Baldi and Peter Sadowski. The dropout learning algorithm. *Artificial intelligence*, 210:78–122, 2014. 2

[3] Jia Deng, Wei Dong, Richard Socher, Li-Jia Li, Kai Li, and Li Fei-Fei. Imagenet: A large-scale hierarchical image database. In *CVPR*, pages 248–255. Ieee, 2009. 4

[4] Felix Elwert and Christopher Winship. Endogenous selection bias: The problem of conditioning on a collider variable. *Annual review of sociology*, 40:31–53, 2014. 3

[5] Ranjay Krishna, Yuke Zhu, Oliver Groth, Justin Johnson, Kenji Hata, Joshua Kravitz, Stephanie Chen, Yannis Kalantidis, Li-Jia Li, David A Shamma, et al. Visual genome: Connecting language and vision using crowdsourced dense image annotations. *IJCV*, 123(1):32–73, 2017. 4

[6] David Lopez-Paz, Robert Nishihara, Soumith Chintala, Bernhard Scholkopf, and Léon Bottou. Discovering causal signals in images. In *CVPR*, pages 6979–6987, 2017. 3

[7] Nitish Srivastava, Geoffrey Hinton, Alex Krizhevsky, Ilya Sutskever, and Ruslan Salakhutdinov. Dropout: a simple way to prevent neural networks from overfitting. *JMLR*, 15(1):1929–1958, 2014. 2

[8] Oriol Vinyals, Alexander Toshev, Samy Bengio, and Dumitru Erhan. Show and tell: A neural image caption generator. In *CVPR*, pages 3156–3164, 2015. 2

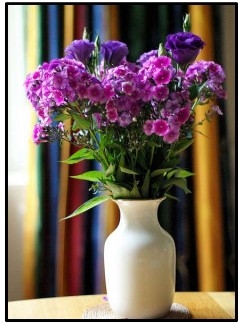
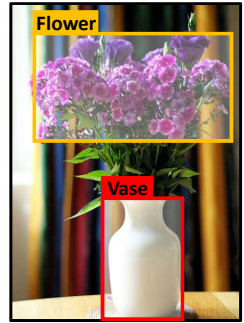

*A vase of flowers sitting on top of a table.*

*A white vase filled with purple flowers on top of a table.*

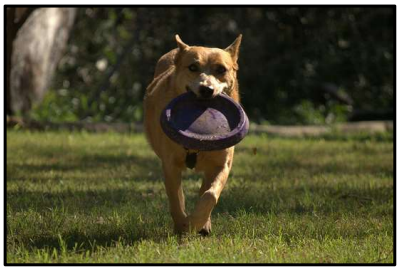
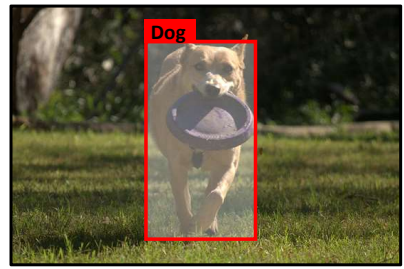

*A dog holding a frisbee in his mouth.*

*A dog is running with a frisbee in his mouth.*

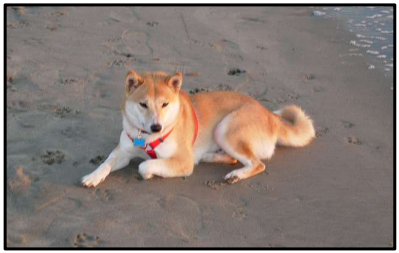
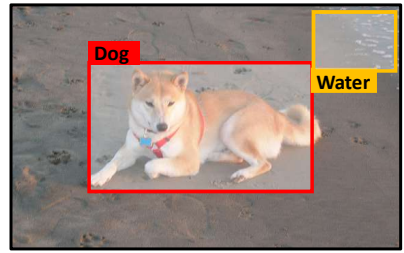

*A couple of dog lying on the beach.*

*A dog lying on the beach next to the water.*

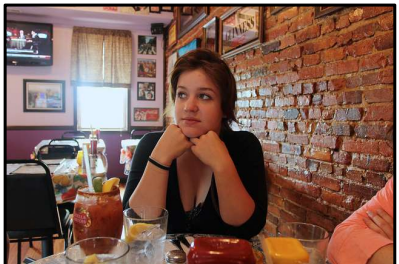
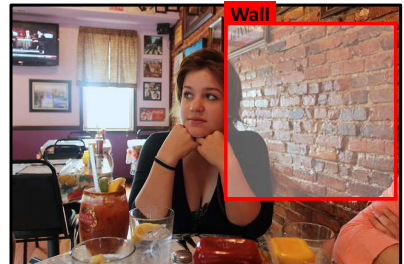

*A woman sitting at a table with a table.*

*A woman sitting at a table in a restaurant.*

Figure 7. Qualitative visualizations in Image Captioning with utilizing Faster R-CNN feature (left) and our VC feature (right). Boxes in image represent the attention region when generating words with the same color.

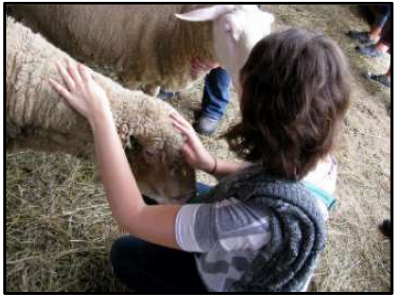
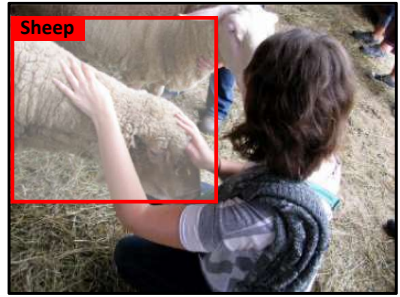

*A girl standing next to a sheep.*

*A women **petting** a sheep in a field.*

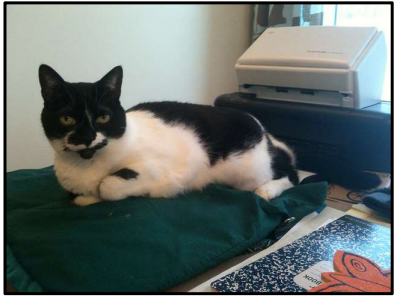
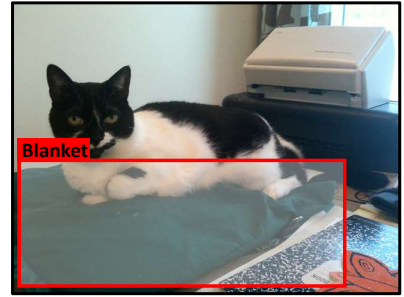

*A black and white cat sitting on a bed.*

*A black and white cat **laying** on a **green blanket**.*

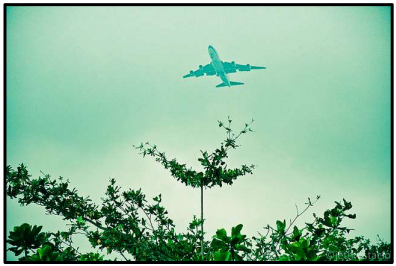
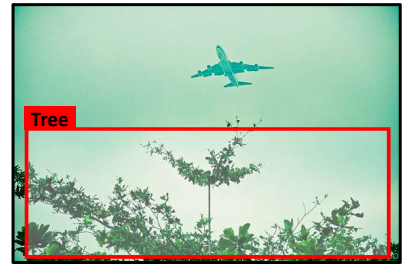

*A plane is flying in the sky.*

*An airplane is flying in the sky **over a tree**.*

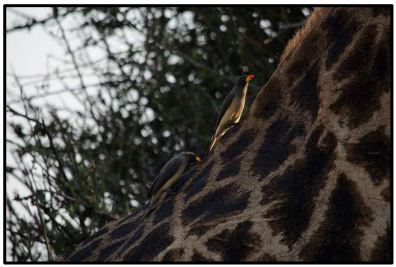
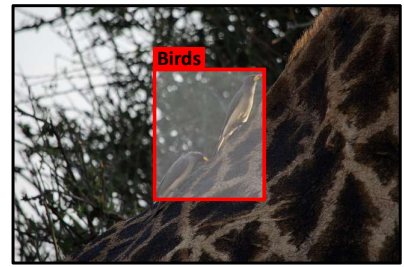

*A bird sitting on top of a tree.*

***Two birds perched** on top of a tree.*

Figure 8. Qualitative visualizations in Image Captioning with utilizing Faster R-CNN feature (left) and our VC feature (right). Boxes in image represent the attention region when generating words with the same color.

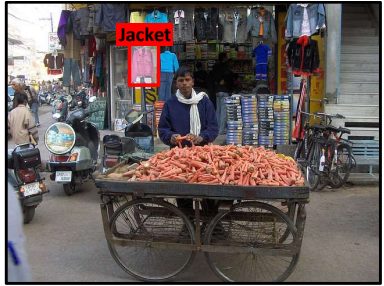 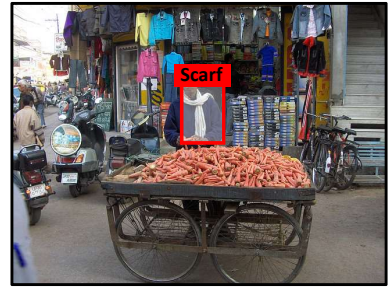

Q: Is the man wearing a scarf?
A: Yes

Q: Is the man wearing a scarf?
A: Yes

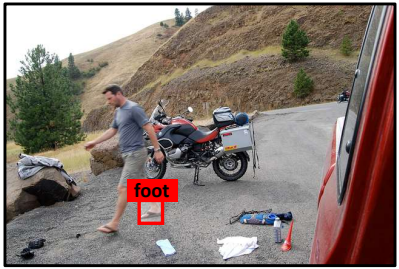 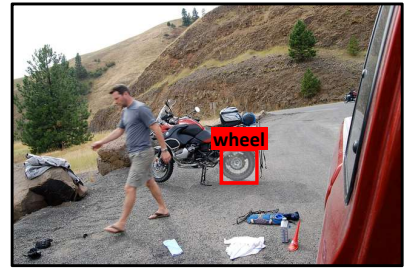

Q: How many wheels does the vehicle behind the man have?
A: 2

Q: How many wheels does the vehicle behind the man have?
A: 2

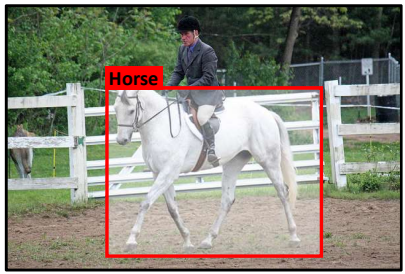 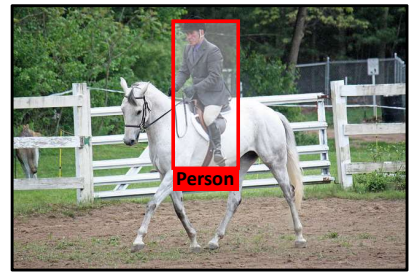

Q: Is the rider a child or an adult?
A: Adult

Q: Is the rider a child or an adult?
A: Adult

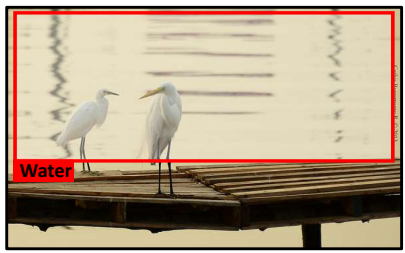 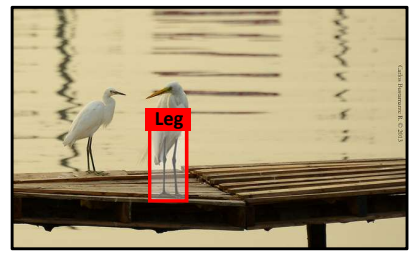

Q: Are the birds legs touching the water?
A: Yes

Q: Are the birds legs touching the water?
A: No

Figure 9. The qualitative results of Visual Question Answering by using the Faster R-CNN feature (left) and concatenated with our VC feature (right). Boxes denote the attended region when answering.

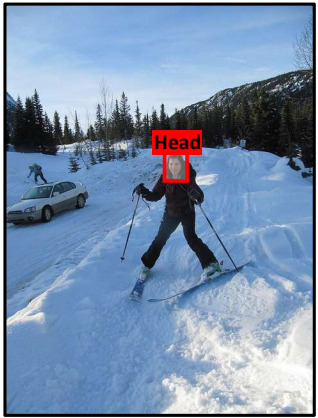

*Q: Is this woman legs stuck?*
*A: No*

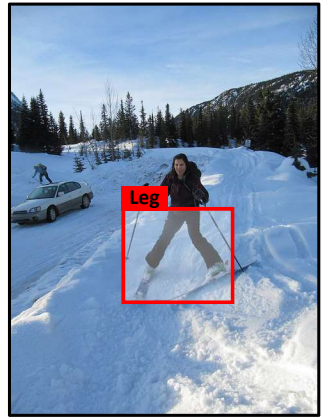

*Q: Is this woman legs stuck?*
*A: No*

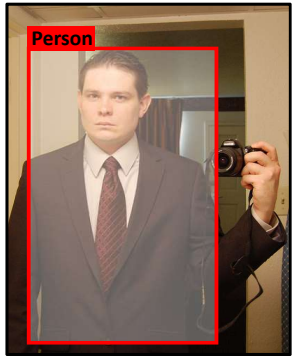

*Q: Is there a camera?*
*A: Yes.*

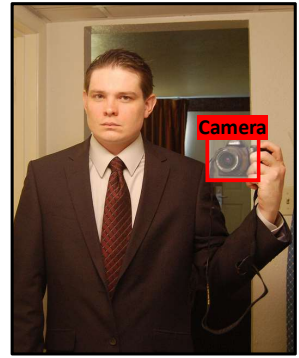

*Q: Is there a camera?*
*A: Yes*

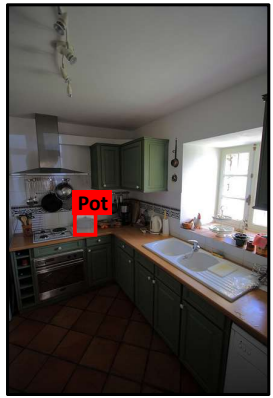

*Q: What is in the sink?*
*A: nothing*

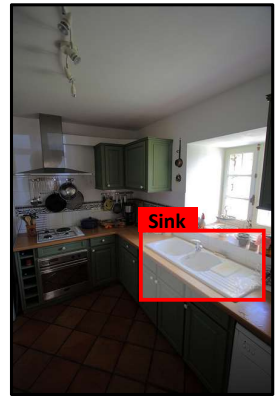

*Q: What is in the sink?*
*A: nothing*

Figure 10. The qualitative results of Visual Question Answering by using the Faster R-CNN feature (left) and concatenated with our VC feature (right). Boxes denote the attended region when answering.



\end{document}